\documentclass[11pt,a4paper]{article}
\usepackage[hyperref]{emnlp-ijcnlp-2019}
\usepackage{times}
\usepackage{latexsym}
\usepackage{graphicx}
\usepackage{subcaption}
\usepackage{multirow}
\usepackage{xcolor} 

\usepackage{url}

\aclfinalcopy 

\definecolor{gred}{RGB}{219,68,55}
\definecolor{gblue}{RGB}{66,133,244}
\definecolor{gyellow}{RGB}{244,180,0}
\definecolor{ggreen}{RGB}{15,157,88}
\definecolor{ggrey}{RGB}{115,115,115}

\newcommand{\nouncol}[2]{\fboxsep1pt \colorbox{ggreen!#2}{\strut #1}} 
\newcommand{\verbcol}[2]{\fboxsep1pt \colorbox{gyellow!#2}{\strut #1}} 

\newcommand{\comment}[1]{}  

\newcommand{\impcom}[1]{}  

\title{Do Massively Pretrained Language Models Make Better Storytellers?}

\author{Abigail See, Aneesh Pappu\thanks{\enspace equal contribution}, Rohun Saxena\footnotemark[1], Akhila Yerukola\footnotemark[1], Christopher D. Manning \\
Stanford University \\
{\tt \string{abisee,apappu,rohun,akhilay,manning\string}@cs.stanford.edu}}

\date{}

\begin{document}
\maketitle
\begin{abstract}
Large neural language models trained on massive amounts of text have emerged as a formidable strategy for Natural Language Understanding tasks.
However, the strength of these models as Natural Language \textit{Generators} is less clear. 
Though anecdotal evidence suggests that these models generate better quality text, there has been no detailed study characterizing their generation abilities.
In this work, we compare the performance of an extensively pretrained model, OpenAI GPT2-117 \cite{radford2019language}, to a state-of-the-art neural story generation model \cite{fan2018hierarchical}.
By evaluating the generated text across a wide variety of automatic metrics, we characterize the ways in which pretrained models do, and do not, make better storytellers.
We find that although GPT2-117 conditions more strongly on context, is more sensitive to ordering of events, and uses more unusual words, it is just as likely to produce repetitive and under-diverse text when using likelihood-maximizing decoding algorithms.
\end{abstract}

\begin{section}{Introduction}
\label{sec:intro}
In 2018, large-scale neural models such as ELMo \cite{peters2018deep}, BERT \cite{devlin2019bert} and OpenAI GPT \cite{radford2018gpt1} emerged as a dominant approach in NLP.
By pretraining on massive amounts of unlabeled text (often orders of magnitude larger than the the target task's labeled dataset), these models achieve state-of-the-art performance across a variety of Natural Language Understanding benchmarks. 
In particular, the OpenAI GPT2 language model \cite{radford2019language} achieves state-of-the-art performance on several language modeling benchmarks, even in a zero-shot setting.
While GPT2's performance as a language model is undeniable, its performance as a text generator is much less clear.
Though the model has generated certain impressive samples of text -- such as a widely-circulated passage about Ovid's Unicorn \cite{radford2019language} -- there has been no detailed study to formalize these observations.

In this work, we perform an in-depth study of the properties of text generated by GPT2-117 (the smallest version of GPT2) in the context of story generation.
By comparing to a state-of-the-art, specialized-architecture neural story generation model \cite{fan2018hierarchical}, we ask the following questions.
In what ways does a large amount of open-domain pretraining data change the characteristics of generated text? 
In what ways does it make no difference? 
And is a task-specific architecture necessary?

For any probabilistic language model, the generated text is strongly affected by the choice of decoding algorithm -- this is especially true for \textit{open-ended} text generation tasks such as storytelling and chitchat dialogue \cite{kulikov2018importance, holtzman2019curious}.
Nevertheless, most natural language generation papers evaluate only \textit{one} decoding algorithm -- this is often due to the time and expense required for human evaluation.
For example, \citeauthor{fan2018hierarchical} use top-$k$ sampling (a decoding algorithm in which $k$ governs the quality-diversity tradeoff), but only evaluate one value of $k$.
However, evaluating one $k$ gives an incomplete view of the generation system -- several researchers have emphasized the importance of evaluating generation systems over the entire quality-diversity spectrum, rather than a single point on it \cite{caccia_language_gans_falling_short,hashimoto2019unifying}.

In this study, we prioritize evaluating text across the \textit{whole} $k$ spectrum, and measuring \textit{many} different automatic metrics, rather than a few human metrics.
Though the lack of human evaluation limits our ability to measure overall quality \cite{liu2016not, novikova2017we, hashimoto2019unifying}, we are able to produce an objectively defined, richly detailed and reproducible evaluation of the generated text.
To our knowledge, this work is the first comprehensive analysis of the characteristics of GPT2-generated text.
Our study provides insight into the effect of large-scale pretraining on open-ended natural language generation, as well as the effect of $k$ on text generated with top-$k$ sampling.
We hope our results will inform other researchers' choice of models, pretraining schemes, and decoding algorithms -- decisions that can often feel like blind choices.
To enable readers to browse the generated text, conduct their own evaluations, or run our evaluations on their own text, we publicly release our generated stories and evaluation code.\footnote{Code and generated stories available at \url{https://github.com/abisee/story-generation-eval}}
\end{section}

\begin{section}{Background}

\paragraph{WritingPrompts dataset}
WritingPrompts \cite{fan2018hierarchical} is a story generation dataset containing 303,358 human-written (\textit{prompt}, \textit{story}) pairs collected from the /r/WritingPrompts subreddit -- a forum where Reddit users compose short stories inspired by other users' prompts.
An example can be seen at the top of Table \ref{tab:main_example}. 
The mean prompt length is 28.4 words and the mean story length is 734.5 words.
The dataset is 887MB of text in total, contains 200 million story words, and is divided into 90\% train, 5\% validation and 5\% test splits.

\paragraph{The Fusion Model}
The Fusion Model is a state-of-the-art neural story generation architecture trained on the WritingPrompts dataset \cite{fan2018hierarchical}.
It is based on the Convolutional Seq2seq model of \citet{gehring2017convolutional} and aims to improve two aspects of story generation: modeling long-range context and increasing relevance of the story to the prompt.
To achieve the former, the model uses a multi-scale gated self-attention mechanism.
For the latter, the model uses a fusion mechanism \citep{Sriram2018} in which one seq2seq model is trained on the task, then frozen, and a second seq2seq model is trained on the task with access to the first model's hidden states. 
Compared to the Convolutional Seq2seq model and other baselines, the Fusion Model achieves improved perplexity, story-prompt relevance and human preference scores.
The Fusion Model has a vocabulary of 104,960 words, a 3-layer encoder and 8-layer decoder in the first seq2seq model, and a 5-layer encoder and 5-layer decoder in the second model -- in total, 255.4 million parameters.

\paragraph{GPT2-117}
GPT2 \cite{radford2019language} is a large Transformer language model trained on WebText, a diverse corpus of internet text (not publicly released) containing over 8 million documents equalling 40GB of text in total. 
The full-size GPT2 model, which has 1542 million parameters, obtains state-of-the-art results on a variety of language modeling and other Natural Language Understanding benchmarks.
At the time of our experiments, \citeauthor{radford2019language} had only released the smallest of the models, known as GPT2-117.\footnote{Since conducting our experiments, larger models have been publicly released. At the time of writing, the full-size GPT2 model has not been publicly released.}
This model, which we use for our experiments, has 12 layers and 117 million parameters.
Like the full-size GPT2 model, it has a vocabulary of 50,257 byte-pair-encoding (BPE) tokens. 
The BPE encoding allows the model to encode and generate any Unicode string, regardless of preprocessing, tokenization, or vocabulary size.
The model has a context size of 1024, meaning it can process text up to 1024 BPE tokens in length.

\paragraph{Decoding algorithms}
Inspired by Neural Machine Translation, most early attempts at open-ended neural text generation (such as conversational response generation) used the \textit{beam search} decoding algorithm \cite{shang2015neural, serban2016building}.
Like greedy decoding, beam search is a \textit{likelihood-maximizing} decoding algorithm -- given the input sequence $x$, the objective is to find an output sequence $y$ which maximizes $P(y|x)$.
However, researchers have shown that for open-ended generation tasks (including storytelling), beam search produces repetitive, generic and degenerate text \cite{holtzman2019curious}.

More recently, \textit{top-$k$ sampling} has emerged as a primary decoding algorithm for open-ended text generation \cite{fan2018hierarchical, radford2019language}.
In top-$k$ sampling, on each step of the decoder the probability distribution over the vocabulary is truncated to the top $k$ tokens, then re-normalized.
The next token is sampled from the new distribution.
Top-$k$ sampling can be regarded as somewhere between a likelihood maximizing algorithm (when $k=1$; greedy decoding) and an unbiased sampling algorithm (when $k=$ vocabulary size).
\citeauthor{fan2018hierarchical} use top-$k$ sampling (with $k=10$) to generate stories, and \citeauthor{radford2019language} show impressive samples of generated text (primarily from the full-size GPT2 model) for $k=40$.

\end{section} 

\begin{section}{Experimental Details}
\label{sec:exp}

\paragraph{Preprocessing}
\citeauthor{fan2018hierarchical} truncate WritingPrompts stories to 1000 words before training and testing.
Due to the limited context size of GPT2-117, we additionally exclude (\textit{prompt}, \textit{story}) examples that are longer than 1024 BPE tokens when concatenated.
The resulting dataset, which we call WritingPrompts-1024, has 192,364 training, 11,115 validation, and 10,686 test examples.

\paragraph{The Fusion Model}
We use the pretrained version of the Fusion Model, which is available in the Fairseq framework \cite{ott2019fairseq}.
For comparability with GPT2-117, we evaluate the Fusion Model on WritingPrompts-1024 (see Table \ref{table:perplexities}), obtaining perplexities similar to those reported by \citeauthor{fan2018hierarchical} on the full WritingPrompts dataset.

\paragraph{GPT2-117}
In order for the model to condition on prompts and generate stylistically correct stories, we finetune GPT2-117 on WritingPrompts-1024.\footnote{We use the PyTorch re-implementation of GPT2-117 available at \url{https://github.com/huggingface/pytorch-transformers}}
We frame WritingPrompts as a language modeling task, representing the prompt and story as a single sequence separated by a delimiter token.
We finetune the pretrained model until convergence using the default hyperparameters provided in the HuggingFace repository (though we reduce batch size to fit on a single GPU), and use the finetuned model for all further evaluations.

We compute the \textit{word-level} perplexity of the finetuned GPT2-117 on the WritingPrompts-1024 dataset.
That is, we normalize the total negative log probability of the target text by the number of \textit{word}-level (i.e. Fusion Model) tokens, not the number of BPE tokens.
This enables us to compare the perplexities of the two models, despite the tokenization difference \cite{radford2019language}.
The finetuned GPT2-117 obtains a test set word-perplexity of 31.54\footnote{This is similar to other GPT2-117 WritingPrompts finetuning experiments \cite{mao2019emnlp, ziegler2019encoder}.} -- six points lower than the Fusion Model.

\begin{table}[t]
\centering
\begin{tabular}{|l|l|l|}
\hline
\textbf{Model} & \textbf{Valid ppl} & \textbf{Test ppl} \\ \hline
Fusion Model & 37.05 & 37.54 \\ \hline
GPT2-117 & 31.13 & 31.54 \\ \hline
\end{tabular}
\caption{Word-level perplexities on WritingPrompts-1024 for the Fusion Model and finetuned GPT2-117.}
\label{table:perplexities}
\end{table}

\paragraph{Generation settings}
For both models, we generate stories using top-$k$ sampling, obtaining 1000 stories (from 1000 different test set prompts) for several values of $k$ ranging from 1 to vocabulary size.
We use softmax temperature 1.
Like \citeauthor{fan2018hierarchical}, we generate exactly 150-word stories and block the Fusion Model from generating \texttt{<UNK>}.

To obtain human-written stories for comparison, we truncate WritingPrompts-1024 test set stories to 150 words (discarding those shorter than 150 words).
To reduce variance, measurements for human stories are computed over this entire set (rather than just 1000 stories).
\end{section}

\begin{section}{Story-prompt relatedness} 
\label{sec:story-prompt}
Prior research has observed that seq2seq systems frequently produce text that is unrelated to the provided context -- particularly under likelihood-maximizing decoding algorithms such as beam search. 
The issue has inspired multiple explanations \cite{jiang2018sequence} and multiple solutions -- such as alternative training objectives \cite{li2016diversity}, decoding objectives \cite{baheti2018generating, see2019makes}, and architectural changes \cite{fan2018hierarchical}.
In this section, we measure how strongly the models condition on the prompt.

\paragraph{Prompt ranking accuracy}
For both models, we compute \textit{prompt ranking accuracy} \cite{fan2018hierarchical}, which measures the language model's sensitivity to the provided prompt. 
Following the methodology of \citeauthor{fan2018hierarchical}, we randomly select 1000 human-written stories from the test set, and measure the probability (according to the model) of each story conditioned on 10 different prompts -- the true prompt, plus nine randomly selected prompts. 
The prompt ranking accuracy of a model is the percentage of cases in which the model assigns a higher probability to the story under its true prompt than under all of the other nine.
We find that GPT2-117 scores \textbf{80.16\%} on this task, while the Fusion Model scores \textbf{39.8\%}.\footnote{\citet{fan2018hierarchical} report a prompt ranking accuracy of 16.3\% for the Fusion Model. 
We provided the authors with our prompt ranking accuracy code (which was built on top of the authors' code).
The authors indicated that the discrepancy may be due to some code version changes between the time of their original experiments and their code release.} 
Random chance scores 10\%.
This striking result indicates that GPT2-117 conditions on the prompt much more strongly than the Fusion Model.
This is notable, especially because the fusion technique is intended to improve story-prompt relevance.

\paragraph{N-gram similarity} 
For $n=1,2,3$, we measure the percentage of generated $n$-grams that also appear in the prompt. 
For all $n$ and $k$, we find that GPT2-117 has a higher overlap (i.e. copies more from the prompt) than the Fusion Model -- see Figure \ref{fig:copied_ngram} in the Appendix.
Furthermore, for $k<100$, the GPT2-117 overlap is generally much higher than human levels.
Both these phenomena can be seen in Table \ref{tab:main_example}, where, for $k=10$, GPT2-117 copies words such as \textit{queen} more often than both the Fusion Model and the human-written story.

\begin{figure}[t]
    \centering
    \includegraphics[width=\columnwidth]{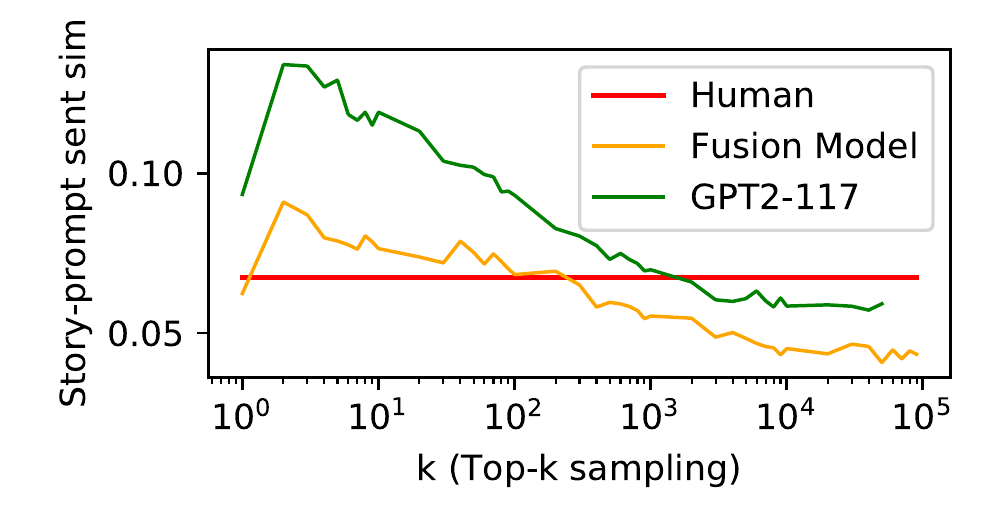}
    \caption{Compared to the Fusion Model, GPT2-117 produces stories that are more semantically similar to the prompt. Similarity decreases as $k$ increases.} 
    \label{fig:sent_sim}
\end{figure}

\paragraph{Sentence embedding similarity}
To capture a higher-level notion of semantic similarity, we measure \textit{story-prompt sentence similarity} -- the cosine similarity of story-prompt sentence pairs, averaged by taking the mean over all pairs.
Sentences are represented by the embedding method of \citet{arora2016simple} -- a weighted average of the GloVe embeddings \cite{pennington2014glove} of the words, with the first principal component removed.
As shown in Figure \ref{fig:sent_sim}, we find a similar pattern as for $n$-gram similarity: GPT2-117 generates sentences that are more similar to the prompt than the Fusion Model for all $k$, and both models' prompt similarity decreases as $k$ increases.

\paragraph{Named entity usage} 
Generally, most named entities mentioned in the prompt (such as \textit{Queen} and \textit{England} in Table \ref{tab:main_example}), should also be mentioned in the story.
Using the spaCy named entity recognizer,\footnote{\url{https://spacy.io}} we measure the \textit{prompt entity usage rate}, which is the percentage of all prompt named entities that appear in the story.\footnote{Given that we limit stories to 150 words, this percentage is lower than it would be if we generated longer stories.}
As shown in Figure \ref{fig:entity-usage} in the Appendix, we find that GPT2-117 uses more of the prompt named entities than the Fusion Model (as well as more named entities overall), but both models use fewer named entities than humans when $k$ is less than vocabulary size.

These patterns can be seen in Table \ref{tab:main_example}: GPT2-117 uses the prompt entities \textit{Queen} and \textit{England} whereas the Fusion Model does not (for either $k$), and GPT2-117 uses specific time entities such as \textit{Thursday} and \textit{3:26 PM}.
While the human story introduces highly-related entities such as \textit{Charles Windsor} and \textit{Prince of Wales} that were not in the prompt, neither model does this (for either $k$).

\paragraph{Conclusion} 
In this section, we found that GPT2-117 conditions on the prompt much more strongly than the Fusion Model -- a result which holds both in language modeling and generation settings.
The latter result supports \citeauthor{radford2019language}'s informal observation that GPT2 has a `chameleon-like' ability to `adapt to the style and content of the conditioning text'.\footnote{\url{https://openai.com/blog/better-language-models/}}
We speculate that GPT2-117's stronger conditioning ability may derive from its Transformer decoder architecture, whose powerful self-attention is used for story-prompt attention.
Though the Fusion Model uses a similar self-attention mechanism in the decoder (i.e., story side), the prompt-story attention has a simpler formulation -- for example, there are no separate key and value vectors \cite{gehring2017convolutional}.
Lastly, we note that very strong prompt-conditioning is not always a good thing -- GPT2-117 often generates stories that copy too much or too literally from the prompt when $k$ is small (this can be seen in Figure \ref{fig:copied_ngram} in the Appendix).

\end{section}

\begin{figure}[t]
    \centering
    \includegraphics[width=\columnwidth]{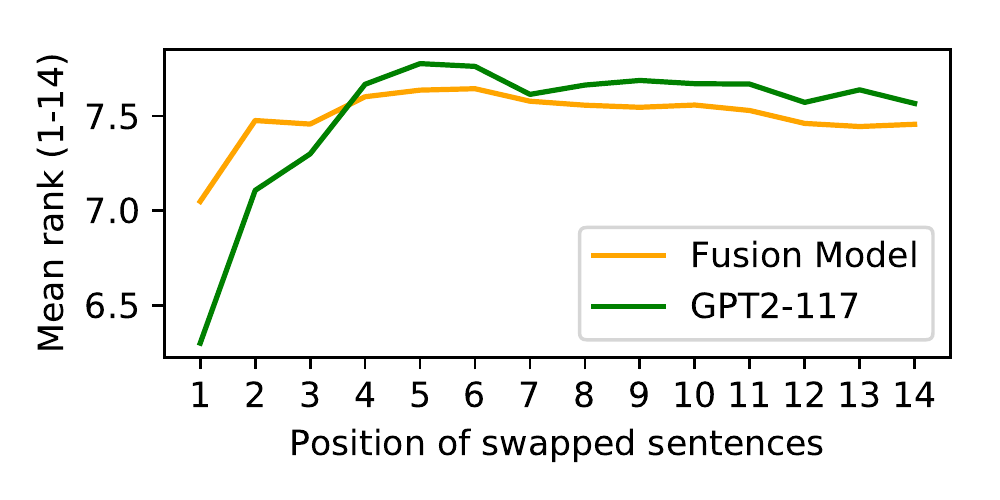}
    \caption{Sensitivity of the models to swapped sentences in different positions.
    A higher mean rank indicates higher sensitivity (i.e. the model assigns lower probability) relative to other positions.
    Both models are less sensitive to swapped sentences at the beginning of the text, compared to later. GPT2-117 shows this pattern more strongly, indicating greater use of context.}
    \label{fig:swap_sensitivity}
\end{figure}

\begin{section}{Coherence}
\label{sec:coherence}
A good story generation model should produce coherent text with a logical ordering of events.
Similarly, the underlying language model should be a good coherence scorer -- assigning higher probability to coherent text than incoherent text.
\citet{Barzilay} evaluate a coherence scorer by measuring its ability to rank shuffled human-written text as less coherent than the original unshuffled text. 
We use this method to evaluate our story generation models.

For each story in the test set, we select the first 15 sentences. 
We then produce 14 corrupted versions of the story by switching each pair of adjacent sentences.
We use the language model to compute the probability of each of the 14 corrupted stories, as well as the original story.
The model's error rate is the percentage of cases in which it rates any of the 14 corrupted candidates better than the original candidate.
Random guessing yields 93.33\% error.
Both models perform well on this task -- the Fusion Model has an error rate of \textbf{3.44\%} and GPT2-117 an error rate of \textbf{2.17\%}.
This 36.92\% error reduction indicates that GPT2-117 is more sensitive to ordering of events.

We also investigate how the position of the swap affects its plausibility (relative to other positions).
Figure \ref{fig:swap_sensitivity} shows, for each swap position, the mean rank assigned to that swap by the model (where rank 1 is the most probable of the 14 corrupted candidates, and rank 14 the least probable).
GPT2-117 assigns a much lower rank to the first few swap positions (i.e., rates them more probable) than the later positions.
The Fusion Model shows a similar but less pronounced pattern.
This shows that both models are less sensitive to out-of-order sentences that occur at the beginning of the text, than those occurring later.\footnote{It's also possible that out-of-order sentences are inherently harder to detect at the beginning of text.}
The stronger pattern for GPT2-117 may be due to its stronger context conditioning (as shown in Section \ref{sec:story-prompt}) -- thus becoming more sensitive as context increases.
However, even for the first three swaps, GPT2-117 is more accurate than the Fusion Model at distinguishing the swapped text from the original.
\end{section}

\begin{section}{Repetition and rareness}
\label{sec:diversity}
Generic, under-diverse and repetitive text is a well-documented problem in neural text generation \cite{jiang2018sequence}. 
While there are many proposed solutions to the problem \cite{li2016mutual,vijayakumar2018diverse,baheti2018generating,zhang2018learning,see2019makes}, it has been shown that a primary cause is likelihood-maximizing decoding algorithms such as greedy decoding and beam search \cite{holtzman2019curious}.
In this section we investigate the role of large-scale pretraining, and the role of $k$, in this problem.

\begin{figure}[t]
    \centering
    \includegraphics[width=\columnwidth]{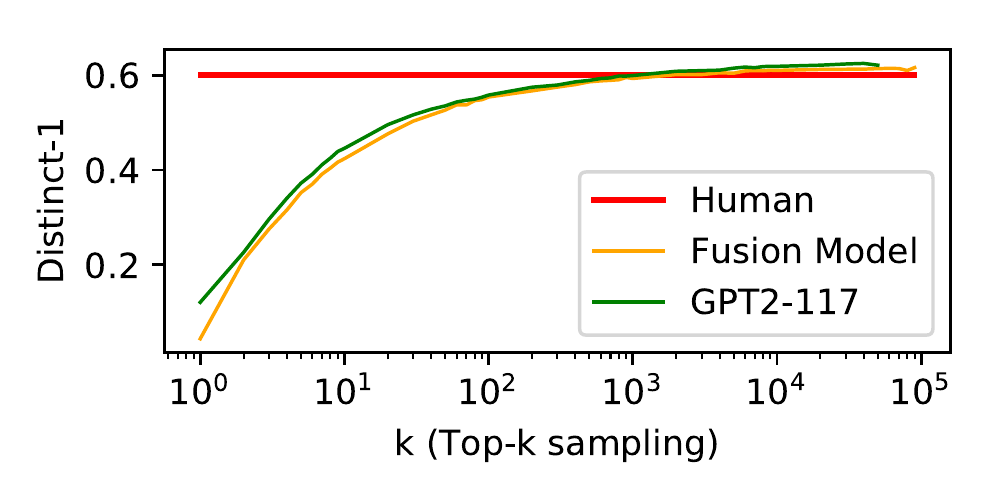}
    \caption{Repetition (low distinct-1) is primarily caused by choice of decoding algorithm (here low $k$), not insufficient training data.
    GPT2-117 is trained on $45 \times$ more data than the Fusion Model, but is similarly repetitive for all $k$.} 
    \label{fig:unique_1gram_mainpaper}
\end{figure}

\paragraph{N-gram repetition} 
The \textit{distinct-$n$} metric of a piece of text is the number of unique $n$-grams divided by the total number of generated $n$-grams \cite{li2016diversity}.
We measure distinct-$n$ of the generated stories for $n=1,2,3$.
A high ratio indicates a high level of within-story lexical diversity, while a low ratio indicates a large amount of within-story repetition.
As shown in Figure \ref{fig:unique_1gram_mainpaper}, both models' unigram diversity is far below that of human text when $k$ is small. 
For example, at $k=10$ (the setting used by \citeauthor{fan2018hierarchical}), the Fusion Model obtains a distinct-1 of $42.4\%$; much less than the human level of $60.0\%$.
This results in a high level of repetition, as shown in Table \ref{tab:main_example}: for $k=10$, both models repeat many phrases (such as \textit{always}, \textit{so scared}, and \textit{finally}).

For bigrams and trigrams, the pattern is similar to unigrams (see Figure \ref{fig:unique_ngram_ratio} in the Appendix).
For both models, distinct-$n$ increases as $k$ increases, converging to a value close to the human level as $k$ approaches vocabulary size.
Though GPT2-117 has a slightly higher distinct-$n$ than the Fusion Model for most values of $k$, the difference is negligible compared to the influence of $k$.
We make three conclusions from these patterns:
(1) Our findings support \citeauthor{holtzman2019curious}'s observation that repetition is strongly related to choice of decoding algorithm, and that likelihood maximizing algorithms (such as top-$k$ sampling with low $k$) are a primary cause of repetition.
(2) The models have in fact learned the correct rate of repetition in human text -- they are able to match this rate when they sample from their full (untruncated) distribution.
(3) Repetition is unlikely to be solved by more pretraining data alone -- even though GPT2-117 is trained on 45 times as much data as the Fusion Model, it produces text that is almost equally repetitive (for equal $k$).

\paragraph{Rare word usage} 
We compute the \textit{mean log unigram probability} of the words in the generated story\footnote{The unigram probability distribution was calculated with respect to the WritingPrompts training set.} -- a high value indicates using fewer rare words while a low value indicates more rare words.
As shown in Figure \ref{fig:rareness_metrics} in the Appendix, word rareness is primarily governed by $k$ -- however, GPT2-117 has a lower mean log unigram probability (i.e., uses more rare words) than the Fusion Model for all equal values of $k \ge 2$.
This can be seen for example in Table \ref{tab:main_example}, where GPT2-117 generates rarer words such as \textit{idle} and \textit{copious} for $k=1000$.
GPT2-117 also generates fewer stopwords than the Fusion Model, for all equal $k$.

GPT2-117's slightly higher rare word usage (compared to the Fusion Model) might be explained by: 
(1) its BPE encoding, which allows it to generate new words, not just those in a fixed vocabulary;
(2) pretraining on a large amount of diverse text, allowing it to learn to produce a greater variety of words;
(3) stronger conditioning on the prompt as described in Section \ref{sec:story-prompt} -- which may inject more rareness into the generated text.

\paragraph{Conclusion} 
Choice of decoding algorithm is a primary factor in diversity and repetition problems, with likelihood-maximizing algorithms the main culprit.
Although GPT2-117 generates more rare words and is very slightly less repetitive than the Fusion Model, the difference is small compared to the effect of $k$, indicating that training data alone is unlikely to solve these problems.
\end{section}

\begin{figure*}[h]
    \centering
    \captionsetup[subfigure]{width=\textwidth,justification=raggedright}
    \begin{subfigure}[t]{0.3\textwidth}
        \centering
        \includegraphics[height=1.15in]{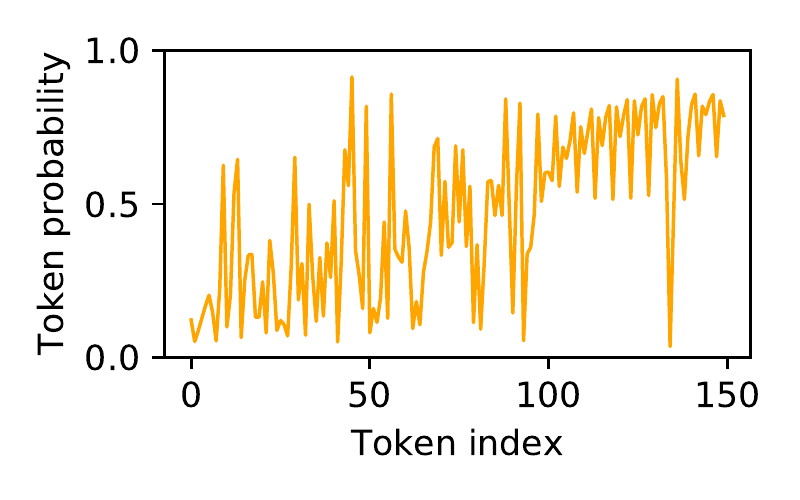}
        \caption{\textbf{Fusion Model} ($k=2$): \textit{I had never seen a man so young before. I had never seen him before, but he had always seemed to be a man of a man. He was young, and he was young. He was a man of a man, and a man who was young, and a man who was} [...]}
    \end{subfigure}
    \hspace{1em}
    \begin{subfigure}[t]{0.3\textwidth}
        \centering 
        \includegraphics[height=1.15in]{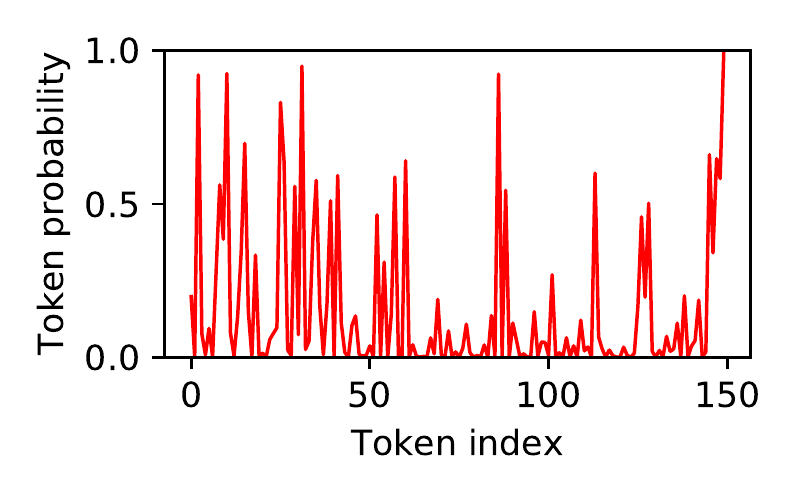}
        \caption{\textbf{Human Text}: \textit{``Looks like the rain's stopped.'' I peered out the window. Art was right; time to get to work. ``Alright, let's move out.'' I could hear the scraping of the stone armor as the men slowly stood. Despite the training,} [...]} 
    \end{subfigure}
    \hspace{1em}
    \begin{subfigure}[t]{0.3\textwidth}
        \centering
        \includegraphics[height=1.15in]{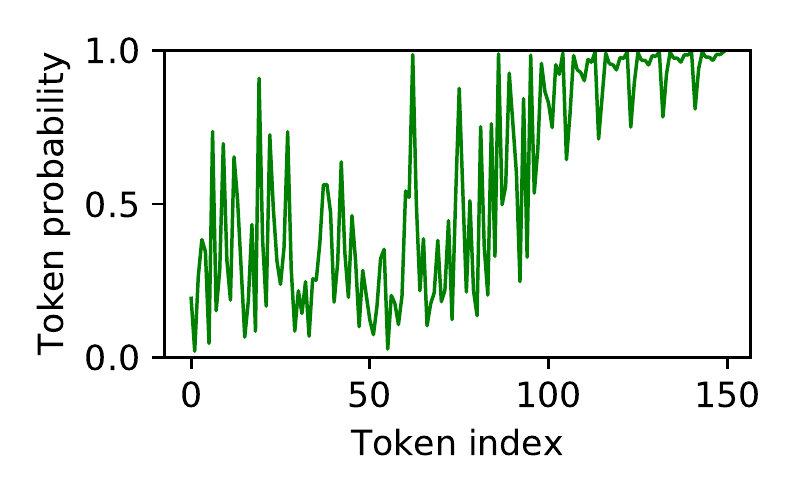}
        \caption{\textbf{GPT2-117} ($k=2$): \textit{I've always been a man of the people. I've always been a strong man. I've always been a strong man. I was born in the city, I was raised in the country. I was raised in a family that wasn't very good. I 'm not a good man.} [...]} 
    \end{subfigure}
    \caption{Under top-$k$ sampling with small $k$ ($k=2$), the two models (left and right) produce text that falls into increasingly confident repeating loops. 
    By contrast, human text (center) maintains an irregular pattern of surprising (low probability) tokens.
    The human text probabilities are measured with respect to the Fusion Model, but similar patterns hold for GPT2-117. 
    Inspired by \citealt{holtzman2019curious}'s figure showing probabilities under beam search.}
    \label{fig:prob_examples}
\end{figure*}

\begin{section}{Syntactic style and complexity}
\label{sec:syn}
A well-trained story generation model should match both the syntactic style and complexity of its training data. 
Low complexity can be a sign of less sophisticated writing, while high complexity can be a sign of poor readability \cite{beers2009syntactic, mcnamara2010linguistic}.
In this section, we measure some features related to the syntactic style and complexity of the generated stories.

\paragraph{Sentence length}
Sentence length is a simple but effective feature to estimate readability and syntactic complexity of text \citep{kincaid1975derivation, roemmele2017evaluating}.
We find that both models generate sentences that are on average shorter than human sentences when $k$ is small, but converge to approximately human length as $k$ increases (see Figure \ref{fig:sent_len} in the Appendix).

\paragraph{Part-of-speech usage}
It has been shown that the distribution of parts-of-speech (POS), and more generally the distribution of POS $n$-grams\footnote{For example, the sentence \textit{I like cats} has the POS bigrams PRONOUN VERB and VERB NOUN.} is a useful feature to represent the style of a piece of text \cite{argamon1998routing, ireland2010language, roemmele2017evaluating}.

Firstly, we compare the part-of-speech distributions of the model-generated text and the human text (see Figure \ref{fig:postag_usage} in the Appendix).
Both models (especially GPT2-117) closely fit the human POS distribution as $k$ approaches vocabulary size.\footnote{One exception is Proper Noun: both models fail to produce enough of these even as $k$ approaches vocabulary size.}
This implies that, as with lexical diversity, the models have no difficulty fitting the statistical distribution of human syntax.
However, under likelihood-maximizing decoding algorithms such as low $k$, a completely different distribution emerges, in which text contains more verbs and pronouns than human text, and fewer nouns, adjectives and proper nouns.

Secondly, we measure the syntactic diversity of the text using the distinct-$n$ metric for POS $n$-grams ($n=1,2,3$) -- see Figure \ref{fig:postag_distinctn} in the Appendix. 
As with lexical diversity (see Section \ref{sec:diversity}), we find that syntactic diversity is similar for the two models, is very low when $k$ is small, and matches human level as $k$ approaches vocabulary size. 
It's likely that for low $k$, the syntactic under-diversity of the text is largely caused by lexical under-diversity (i.e. repetition).
However, we note that as $k$ increases, lexical diversity reaches human level sooner than syntactic diversity -- for example, GPT2-117's lexical distinct-3 reaches human level at $k=600$ (Figure \ref{fig:unique_3gram_ratio}), but its POS distinct-3 reaches human level at $k=6000$ (Figure \ref{fig:postag_distinct3}). 
This implies that, even when the text is no more repetitive than human text, it may still be syntactically repetitive (using the same part-of-speech patterns repeatedly).

\paragraph{Conclusion}
We find when $k$ is small, syntactic complexity of generated text is low, consisting of shorter sentences and a narrower range of syntactic patterns. 
However, as $k$ approaches vocabulary size, the syntactic style of generated text closely matches human syntactic patterns.
As with $n$-gram diversity in Section \ref{sec:diversity}, our results show that syntactic under-diversity is primarily caused by low $k$, not insufficient training data.
\end{section}

\begin{section}{The element of surprise}
\label{sec:surprise}

\paragraph{Model confidence over time}
Several researchers have observed that \textit{model over-confidence} (the model placing high probability on a small range of tokens) can cause poor quality generation \cite{jiang2018sequence,holtzman2019curious}.
In particular, they show that for likelihood-maximizing decoding algorithms such as beam search, model confidence can increase in a snowball-like effect, getting stuck in a loop of repetitive but increasingly self-confident text.
We observe this problem in both our models when $k$ is small.
For example, in Figure \ref{fig:prob_examples}, both models fall into self-reinforcing repetitive loops with rising confidence.
The loop is difficult to break -- the Fusion Model briefly escapes (shown as a sudden downwards spike), but quickly returns.
By contrast, the human text does not show a strong rising trend in probability, and intermittently uses low probability words throughout.\footnote{\citet{gehrmann2019gltr} also identify presence of low probability words as an indicator of human-generated text.}

We formalize these anecdotal observations by measuring the average probability of each of the first 150 word-level tokens in the story (Figure \ref{fig:increasing_prob}).
We find that even when teacher-forcing on human text, the token probabilities increase slightly as the story progresses.
This is likely due to the usefulness of additional context, which increases the model's prediction accuracy.
By comparison, we find that when generating with top-$k$ sampling, the probabilities increase more rapidly, and the increase is even more rapid for smaller $k$.
This confirms that likelihood-maximizing decoding algorithms (such as top-$k$ sampling with small $k$) lead to more rapidly increasing model over-confidence.
Furthermore, we find this pattern holds for both models, with probabilities increasing at a similar rate for equal $k$.
This indicates that, like repetition, model over-confidence is unlikely to be solved by more training data, and is largely governed by choice of $k$.

\begin{figure}[t]
    \centering
    \includegraphics[width=\columnwidth]{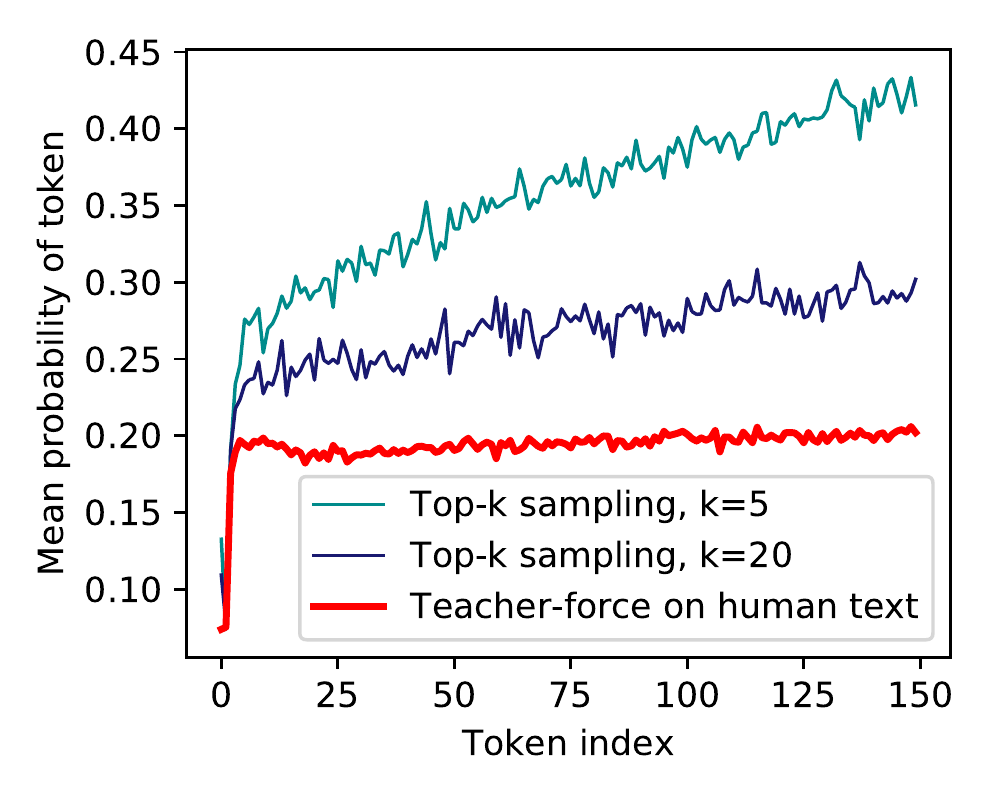}
    \caption{Mean probability for each of the first 150 word-level story tokens.
    When teacher-forcing the model on human text, probability increases slowly.
    When generating with top-$k$ sampling, probability increases faster, especially for smaller $k$. 
    This plot is for the Fusion Model; similar patterns hold for GPT2-117.} 
    \label{fig:increasing_prob}
\end{figure}

\paragraph{Overall model confidence}
We also measure the models' overall confidence, as represented by the total log probability (according to the model) of the generated story.
For both models, we find that story probability decreases as $k$ increases -- see Figure \ref{fig:story_logprob} in the Appendix.
This makes sense, as higher $k$ means sampling tokens with lower probability.
As $k$ approaches the vocabulary size, the Fusion Model's generated story probability matches the probability it assigns to human-written WritingPrompts stories.
Interestingly however, the same is not true for GPT2-117, which converges to a story probability that is \textit{lower} than the probability it assigns the human stories.
This means that under full (non-truncated) sampling, the Fusion Model produces text that is \textit{equally surprising} (to itself) as the WritingPrompts stories, whereas GPT2-117 produces text that is \textit{more surprising} to itself.
Explaining this observation is an open question -- we speculate that GPT2-117's WebText pretraining may cause it to generate (under high $k$) text in a style or genre that is less predictable than WritingPrompts stories.
\end{section}

\begin{table*}[ht]
\footnotesize
\centering
\begin{tabular}{ll}

    \hline \\ [-1.5ex] 
    
    \multicolumn{2}{l}{
        \begin{minipage}{6in} \begin{flushleft}  
            \textbf{Prompt:} You just discovered England's biggest secret: every time they say long live the queen it extends the queens life. 
        \end{flushleft} \end{minipage}
        } 
        
    \\ \\ [-2ex] \hline \\ [-1.5ex]
    
    \multicolumn{2}{l}{
        \begin{minipage}{6in} \begin{flushleft}  
            \textbf{Human-Written Story:} I am immortal, I have inside me blood of kings. My name in Charles Windsor and I am heir to a throne. Prince of Wales, they've called me that my entire life and now I lay on my death bed and my mother - still going strong. I'm not sure how it's possible but she seems ageless, almost timeless. My whole life I've waited for that [...]
        \end{flushleft}\end{minipage}
        } 
        
    \\ \\ [-1ex] \hline \\ [-1.5ex] 
    
    \begin{minipage}{3in} \begin{flushleft}  
        \textbf{Fusion Model ($k=10$)}: \textit{I have no idea when my father would come to visit this place, I was always so excited. I was always so scared. I had to go back to work. I was a little scared of him, but when I was younger, he had never come in for a visit. His mother told me, ``I'm so scared'', as he always did. She never talked to me about anything} [...]
    \end{flushleft} \end{minipage}
    & 
    \begin{minipage}{3in} \begin{flushleft}  
        \textbf{GPT2-117 ($k=10$)}: \textit{* '' So what are we going to do? `` * It was the first time she asked me this question that I ever had asked her. It has taken years but finally I am finally able to do it. It seems that this queen is the queen of England and she's finally the one to tell me that this queen is the reason why she's not being allowed to die.} [...]
    \end{flushleft} \end{minipage}
        
    \\ \\ [-1ex] \hline \\ [-1.5ex] 

    \begin{minipage}{3in} \begin{flushleft}  
        \textbf{Fusion Model ($k=1000$)}: \textit{``Where am I? What happened?'' ``Having been working on my job for over 6 hours now, I do not know how you worked!'' ``I have been working for the last three years. Surely I am an ancient god now.'' The bar patrons snickered. ``Hello?'' ``Those last three years have been worse than a year ago.'' Pain.} [...]
    \end{flushleft} \end{minipage}
    & 
    \begin{minipage}{3in} \begin{flushleft}  
        \textbf{GPT2-117 ($k=1000$)}: \textit{It was an odd occasion for the Queen of England to meet with her. The plane sat idle at 3:26 PM on a Thursday night. Yesterday, the Queen had taken it upon herself to try and get a good look at the plane which had recently been found abandoned. A copious amount of curious glances from around the room until} [...]
    \end{flushleft} \end{minipage}
    
    \\ \\ [-1ex] \hline 
    
\end{tabular}
\caption{A prompt and human story from the dataset, plus the models' top-$k$ generated stories, for two values of $k$.}
\label{tab:main_example}
\end{table*}

\begin{section}{Concreteness}
\label{sec:concrete}
\citet{Brysbaert2014} define the \textit{concreteness} of a word as `the degree to which the concept denoted by a word refers to a perceptible entity'.
Concrete words are generally easier to remember than abstract words, and psycholinguists have theorized they may be learned differently (i.e., concrete words by direct experience and abstract words by text and discourse).
\citeauthor{Brysbaert2014} provide human concreteness ratings for 40,000 common English lemmas rated on a scale from 1 to 5.\footnote{For example, the nouns \textit{television}, \textit{darkness}, and \textit{idea} are rated 4.83, 3.85 and 1.61 respectively, and the verbs \textit{talk}, \textit{see}, and \textit{hope} are rated 4.07, 3.21 and 1.25 respectively.}
We use these ratings to measure the mean concreteness of the nouns and verbs in the story text -- see Figure \ref{fig:concreteness} in the Appendix.

We find that, for the same $k$, GPT2-117 tends to generate more concrete words than the Fusion Model, and that for both models, concreteness converges to approximately human levels as $k$ increases. 
Interestingly, however, when $k$ is small, the noun concreteness is much \textit{higher} than human levels, whereas the verb concreteness is much \textit{lower}. 
This indicates that for small $k$, both models produce stories that, compared to human-written stories, have too many physical objects (as opposed to abstract nouns), and too few physical actions (as opposed to abstract verbs).
This reflects the trend demonstrated in Table \ref{tab:main_example}: when $k$ is small, the models tend to generate descriptive sentences with mostly \textit{is} verbs (e.g. \textit{I was always so excited}), and physical nouns (e.g. \textit{mother, father, queen}).
Only when $k$ increases do we see more tangible actions (e.g. \textit{The bar patrons snickered}) and abstract nouns (e.g. \textit{pain, glances}).
A detailed example, with all nouns and verbs annotated with concreteness, is in the Appendix (Table \ref{fig:concrete_ex}).

\end{section}

\begin{section}{Conclusions}
\label{sec:conclusion}

\paragraph{The effect of massive pretraining}
In this study, we find that GPT2-117 is a better story generation model than the Fusion Model in several specific ways: it conditions much more strongly on the provided context, is more sensitive to correct ordering of events, and generates text that is more contentful (using more rare words, concrete words, and named entities).
In particular, the stronger conditioning result is striking, as the Fusion Model is a complex task-specific architecture designed to increase story-prompt relevance. 
This demonstrates that sometimes, a general-purpose model architecture can outperform a complex task-specific architecture when provided with enough pretraining data.

However, we find that in other aspects, GPT2-117 performs \textit{no better} than the Fusion Model: when $k$ is small, the models generate text that is equally lexically under-diverse, syntactically under-complex, and repetitive -- with a tendency to fall into a snowball effect of increasing over-confidence.
However, these problems correct themselves (i.e., the metrics match human levels) when the models generate from their untruncated distribution.
Our results show that these oft-cited neural generation problems are \textit{not} the fault of the models themselves (which are in fact statistically well-trained to match human text for these metrics), nor caused by too little training data (as these problems are not improved by GPT2-117's extensive pretraining).
Instead, they are primarily caused by likelihood-maximizing decoding algorithms -- such as greedy decoding, beam search, and top-$k$ sampling with low $k$.

\paragraph{The effect of $k$} 
This study detailed the typical characteristics of long-form text generated by neural language models in open-ended settings, under both high entropy (large $k$) and low entropy (small $k$) decoding algorithms.
The negative characteristics of low $k$ output (genericness, repetition, over-simplicity) are by now familiar to researchers.
However, we also uncovered some less obvious characteristics of low-$k$ generated text: compared to human-written text, it tends to copy more from the provided context (particularly GPT2-117); it contains more verbs and pronouns but fewer nouns and adjectives; its nouns are more concrete but its verbs are less concrete; and it uses a smaller range of syntactic patterns (a phenomenon that can't be entirely attributed to $n$-gram repetition).

As $k$ increases to vocabulary size, we find that the model-generated text closely fits the human text on most of the metrics we measured. 
However, it is clear by inspection that the high-$k$ model-generated text lacks many crucial aspects such as commonsense reasoning, world knowledge and multi-sentence coherence -- an example of this superficially fluent but nonsensical text can be seen in Table \ref{tab:weird} in the Appendix.
We believe that true progress in open-ended Natural Language Generation will come from attempting to address these high $k$ problems -- i.e., strategies to imbue the language model with better reasoning, knowledge and planning abilities -- rather than continuing to seek ways to mitigate the diversity and repetition problems of the low $k$ setting.

\paragraph{Limitations of this study}
This study uses only the smallest version of GPT2. 
It is likely that the larger versions of GPT2 may exhibit stronger statistical differences for the metrics we examine.
Such a study would illustrate the effect of larger model capacity, and more fully reveal the possible benefits of massive pretraining.
We release our annotation code so that other researchers may repeat our study on more models and datasets.

This study did not include human evaluation, which is currently the only reliable way to assess overall text quality, as well as quantify the deficiencies of high $k$ output described above (coherence, reasoning, and world knowledge).
As such, this study quantifies the \textit{diversity} side more than the \textit{quality} side of the quality-diversity tradeoff.
Consequently, this study demonstrates the importance of developing better methods to computationally quantify notions such as text coherence, logicality and commonsense correctness -- an effort that may ultimately hold the key to generating text with those desirable attributes.

\end{section}

\begin{section}{Acknowledgments}
This work was funded by the Gerald J. Lieberman Fellowship, Tencent, and the DARPA CwC program under ARO prime contract no. W911NF-15-1-0462.
We also thank the reviewers for their helpful comments.
\end{section}

\clearpage

\bibliography{emnlp-ijcnlp-2019}
\bibliographystyle{acl_natbib}

\onecolumn
\appendix
\section*{\Large Appendix}

\begin{figure*}[h]
\centering
    \begin{subfigure}{0.45\textwidth}
      \centering
      \includegraphics[height=2in]{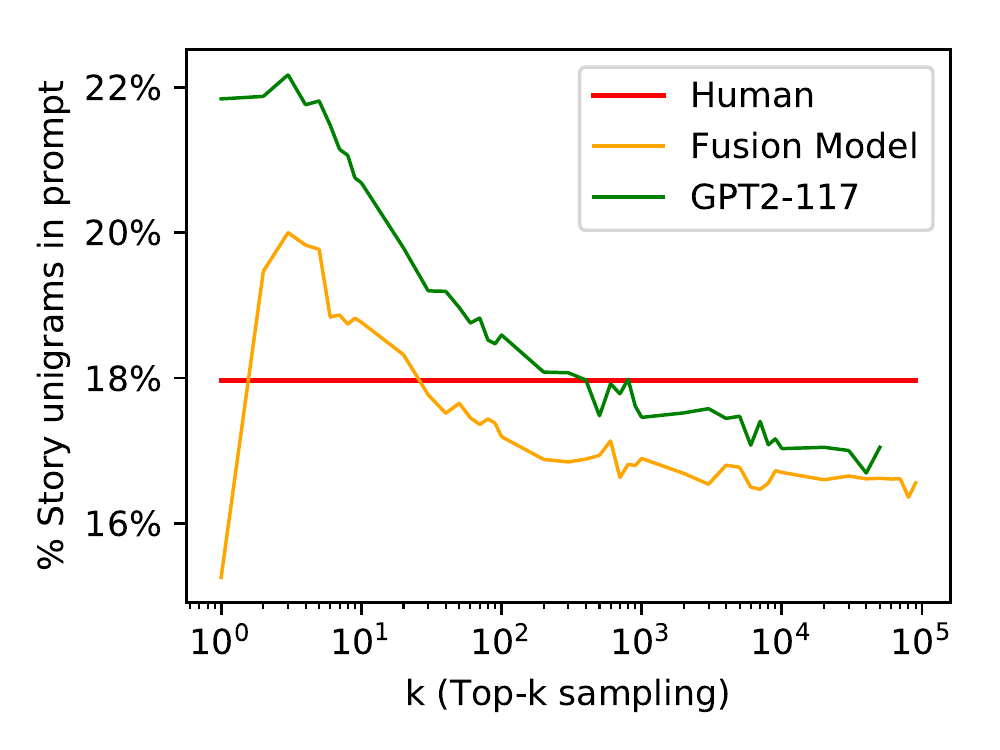}
      \caption{Percent of all story unigrams that are in the prompt.}
      \label{fig:copied_1gram}
    \end{subfigure}
    
    \vspace{1em}
    \begin{subfigure}{0.45\textwidth}
      \centering
      \includegraphics[height=2in]{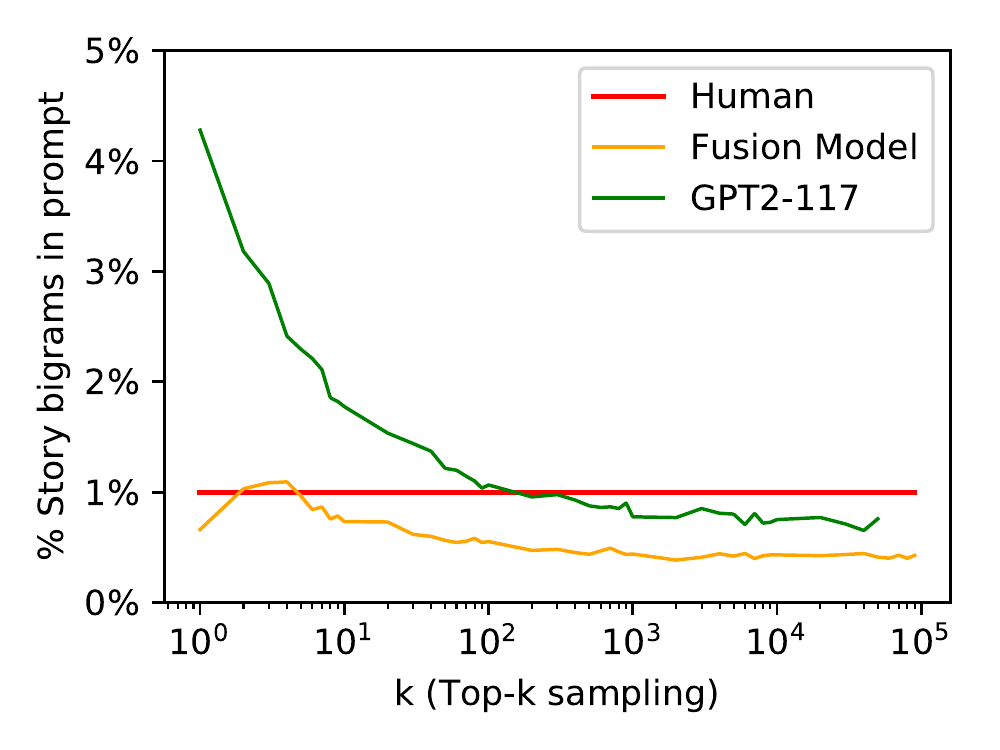}
      \caption{Percent of all story bigrams that are in the prompt.}
      \label{fig:copied_2gram}
    \end{subfigure}
    
    \vspace{1em}
    \begin{subfigure}{0.45\textwidth}
      \centering
      \includegraphics[height=2in]{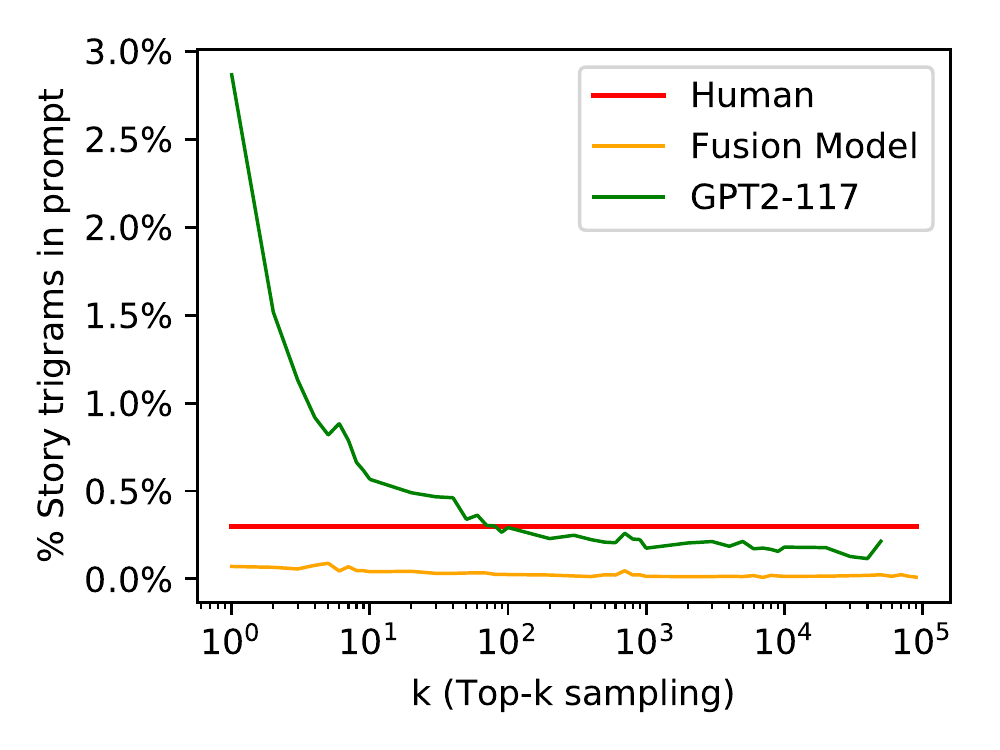}
      \caption{Percent of all story trigrams that are in the prompt.}
      \label{fig:copied_3gram}
    \end{subfigure}
    
\caption{$n$-gram similarity between prompt and story, for $n=1,2,3$, for both models and all $k$. GPT2-117 copies many more $n$-grams from the prompt than the Fusion Model. See Section \ref{sec:story-prompt} for discussion.}
\label{fig:copied_ngram}
\end{figure*}

\clearpage

\begin{figure*}
    \begin{subfigure}{0.45\textwidth}
      \centering
      	\includegraphics[height=2in]{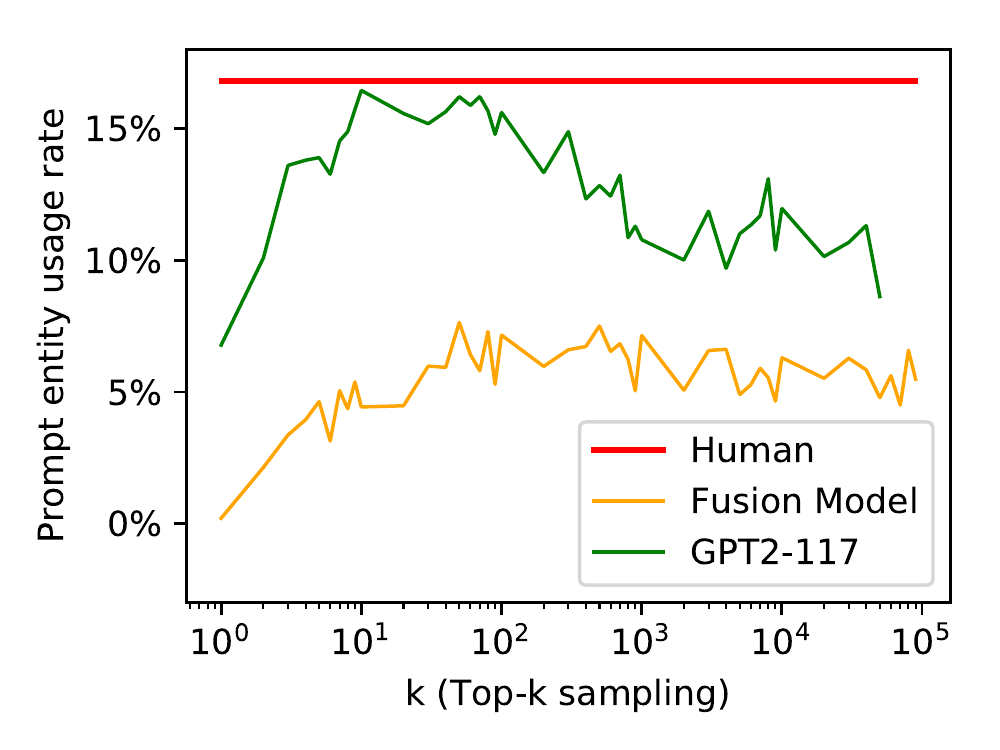}
    	\caption{The proportion of all prompt named entities that are used in the story.}
		\label{figs:p-en-usage}
    \end{subfigure}
    \hspace{2em}
    \begin{subfigure}{0.45\textwidth}
      \centering
      \includegraphics[height=2in]{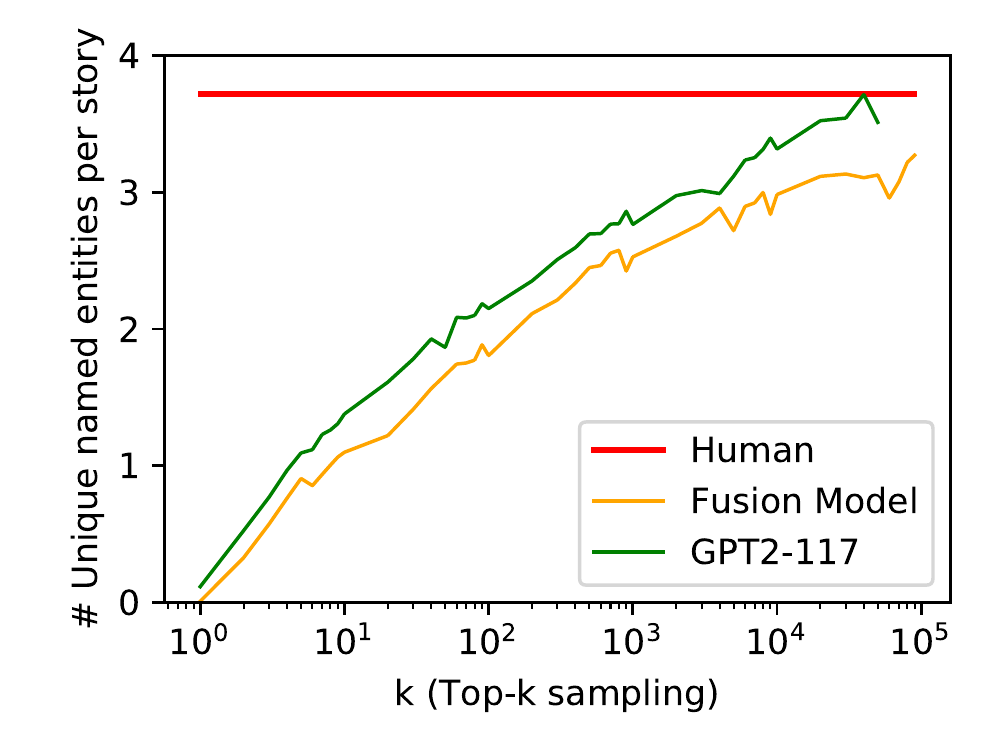}
    	\caption{The number of unique named entities that appear in the story.}
		\label{figs:num-en}
    \end{subfigure}
    
    \vspace{1em}
    
\caption{Prompt entity usage rate (left) and mean number of unique named entities in the story (right), for both models and all $k$. 
GPT2-117 generally uses a larger proportion of the prompt named entities, and more named entities overall, than the Fusion Model. 
Both models generally use fewer named entities than human text when $k$ is less than vocabulary size.
See Section \ref{sec:story-prompt} for discussion.
}
\label{fig:entity-usage}
\end{figure*}

\begin{figure}
    \centering
    \includegraphics[height=2in]{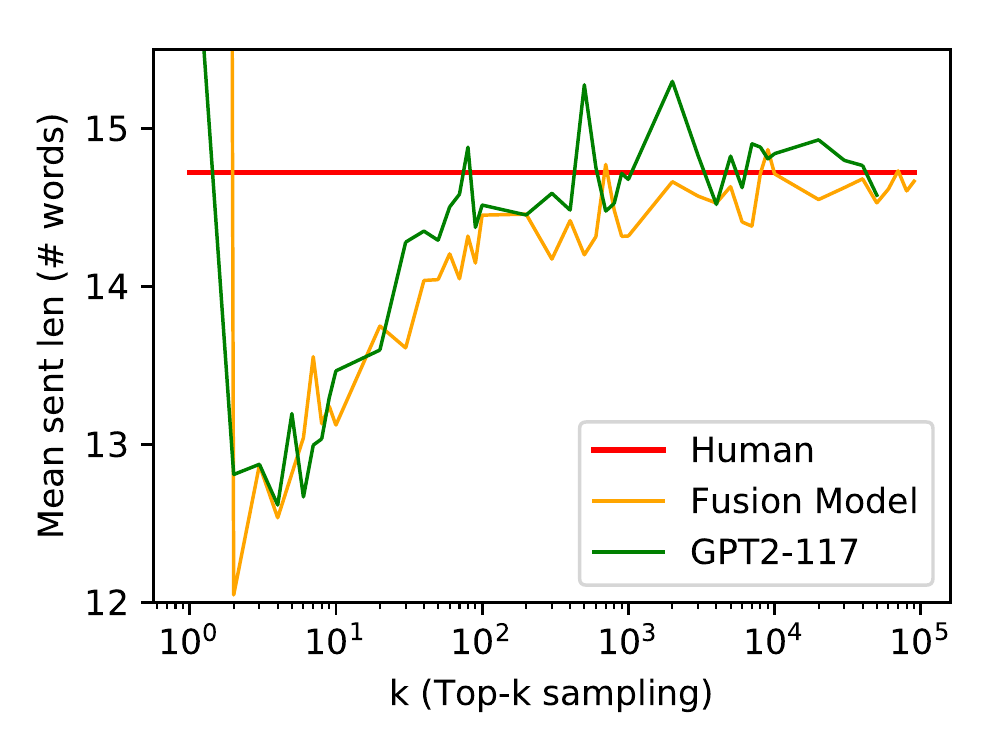}
    \caption{Mean sentence length for both models and all $k$. For both models, sentence length increases as $k$ increases. The spike at $k=1$ is due to long repeating sequences with no sentence-ending token.
    See Section \ref{sec:syn} for discussion.}
    \label{fig:sent_len}
\end{figure}

\clearpage

\begin{figure*}
\centering

    \begin{subfigure}{0.45\textwidth}
      \centering
      \includegraphics[height=2in]{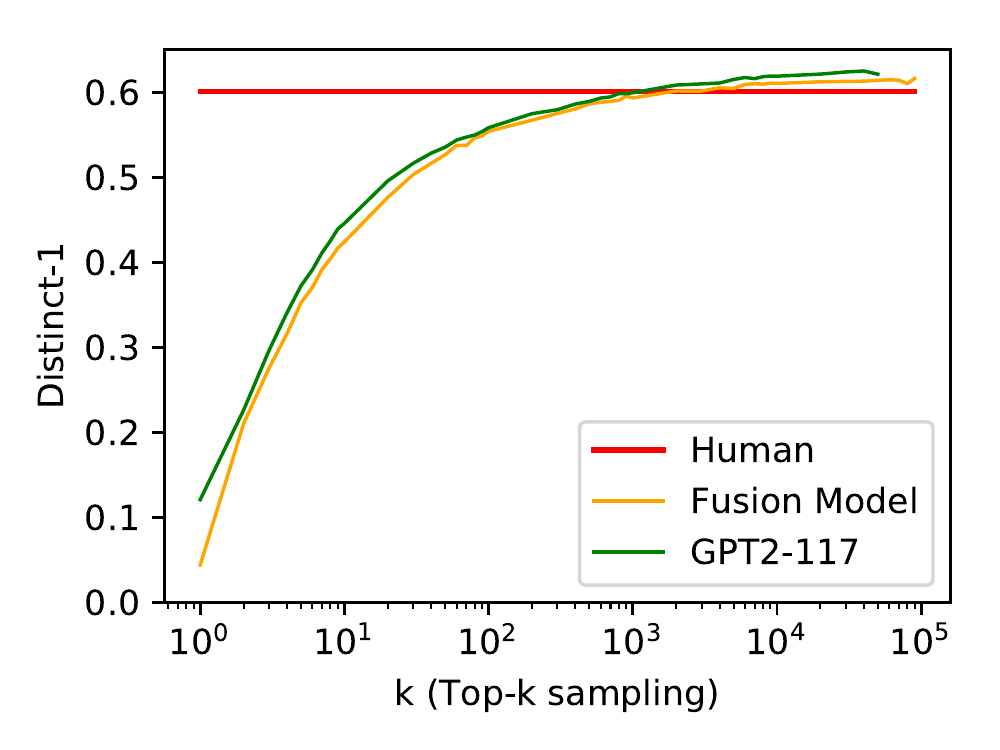}
      \caption{Distinct-1 (ratio of unique unigrams in the story to total number of generated unigrams in the story).}
      \label{fig:unique_1gram_ratio}
    \end{subfigure}
    
    \vspace{1em}
    
    \begin{subfigure}{0.45\textwidth}
      \centering
      \includegraphics[height=2in]{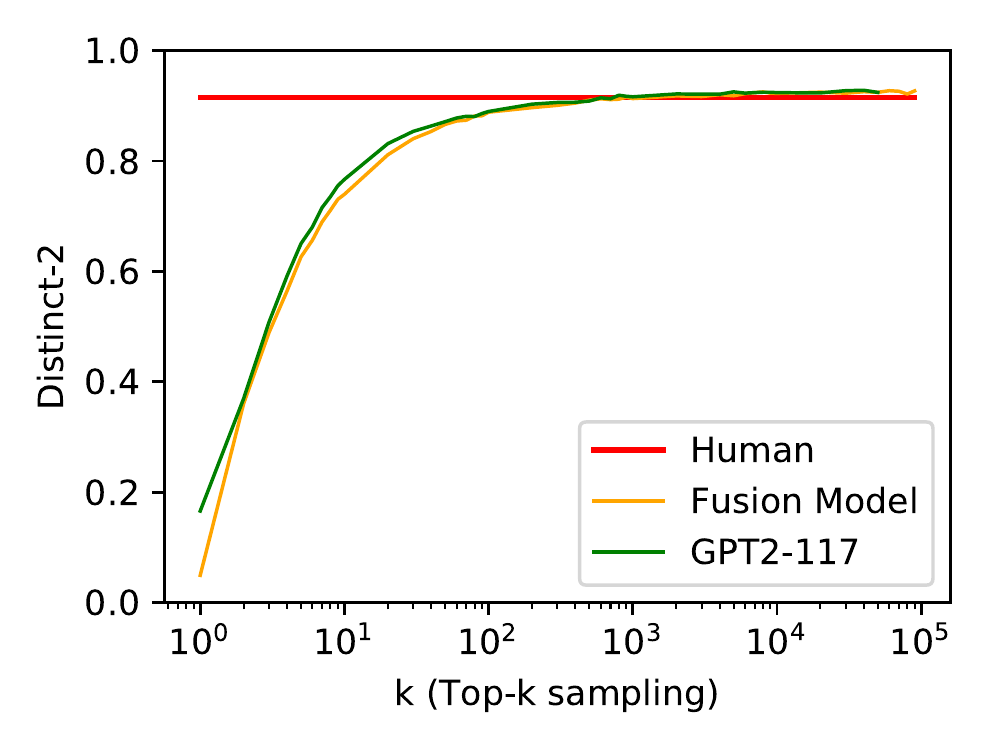}
      \caption{Distinct-2 (ratio of unique bigrams in the story to total number of generated bigrams in the story).}
      \label{fig:unique_2gram_ratio}
    \end{subfigure}
    
    \vspace{1em}
    
    \begin{subfigure}{0.45\textwidth}
      \centering
      \includegraphics[height=2in]{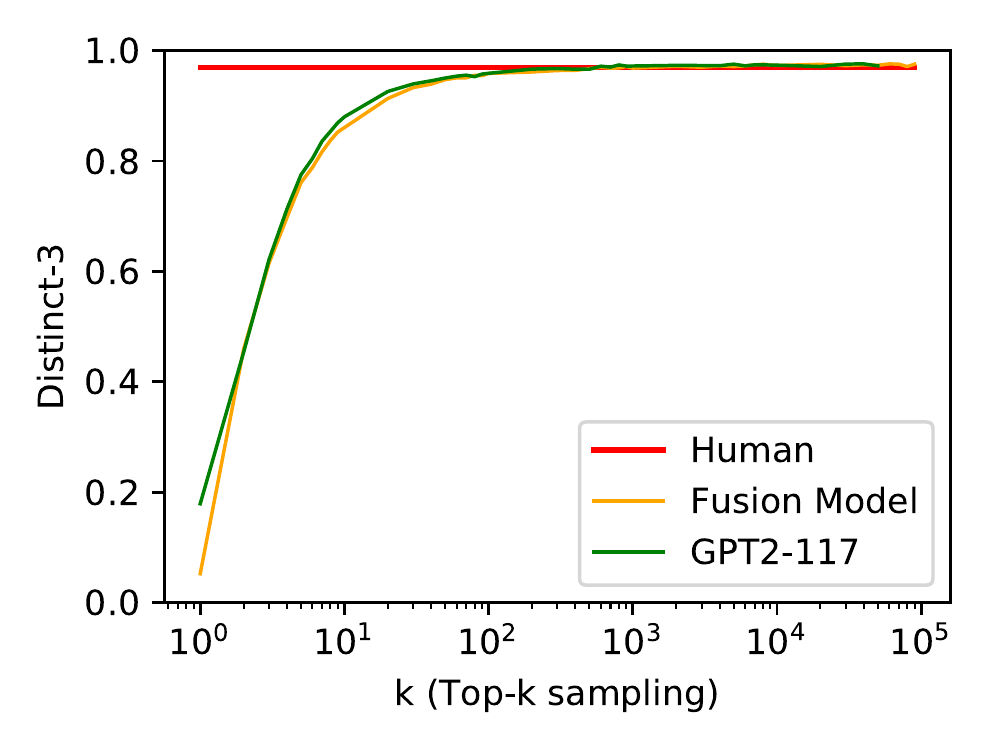}
      \caption{Distinct-3 (ratio of unique trigrams in the story to total number of generated trigrams in the story).}
      \label{fig:unique_3gram_ratio}
    \end{subfigure}
    
    \vspace{1em}
    
\caption{Distinct-$n$ for $n=1,2,3$, for both models and all $k$. 
The ratios, which represent lexical diversity, increase as $k$ increases, with GPT2-117 reaching human levels at $k=2000$ for unigrams, $k=800$ for bigrams and $k=600$ for trigrams.
Lexical diversity is slightly higher for GPT2-117 than for the Fusion Model for equal $k$, but the primary determining factor is $k$.
See Section \ref{sec:diversity} for discussion.}
\label{fig:unique_ngram_ratio}
\end{figure*}

\clearpage

\begin{figure*}
\centering

    \begin{subfigure}{0.45\textwidth}
      \centering
      \includegraphics[height=2in]{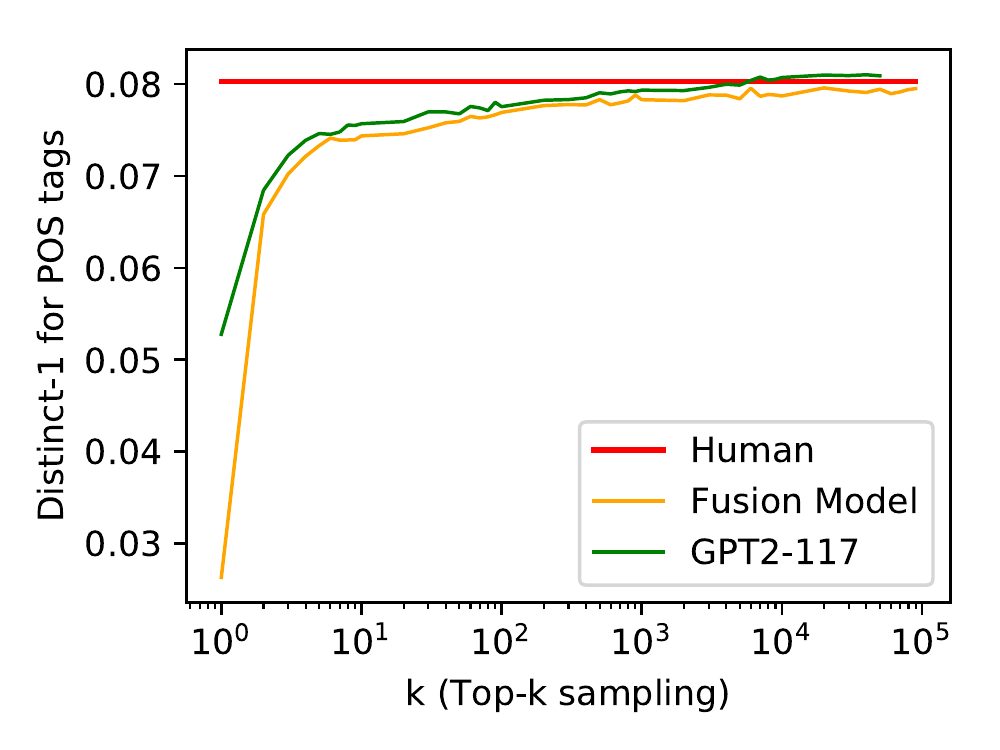}
      \caption{POS tag distinct-1 (ratio of unique POS unigrams in the story to total number of generated POS unigrams in the story).}
      \label{fig:postag_distinct1}
    \end{subfigure}
    
    \vspace{1em}
    
    \begin{subfigure}{0.45\textwidth}
      \centering
      \includegraphics[height=2in]{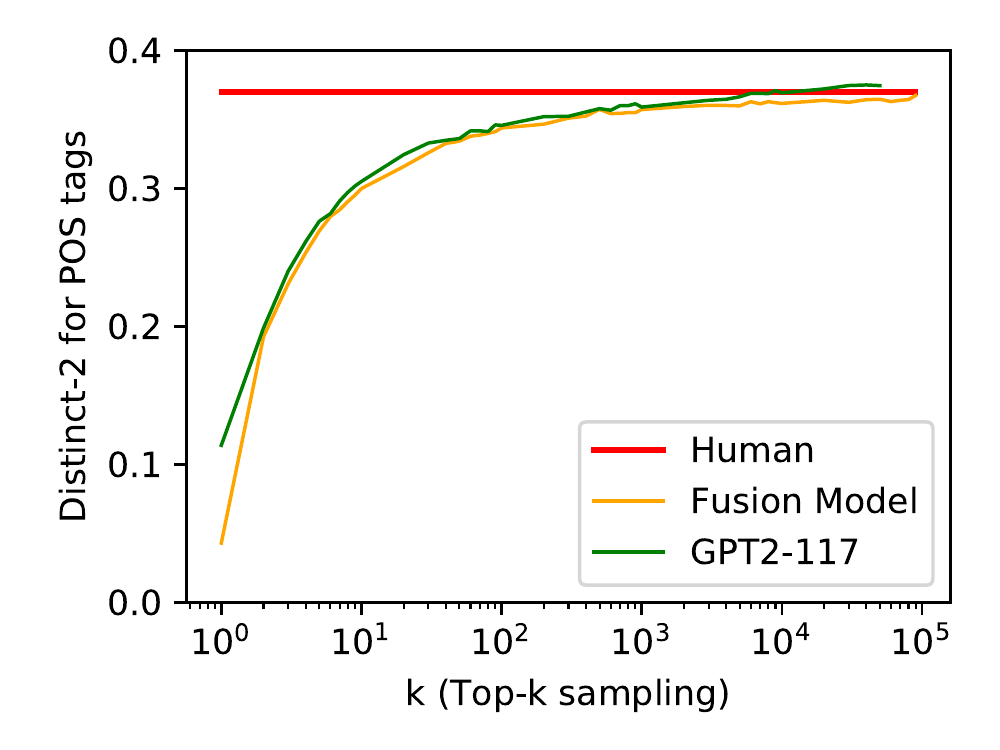}
      \caption{POS tag distinct-2 (ratio of unique POS bigrams in the story to total number of generated POS bigrams in the story).}
      \label{fig:postag_distinct2}
    \end{subfigure}
    
    \vspace{1em}
    
    \begin{subfigure}{0.45\textwidth}
      \centering
      \includegraphics[height=2in]{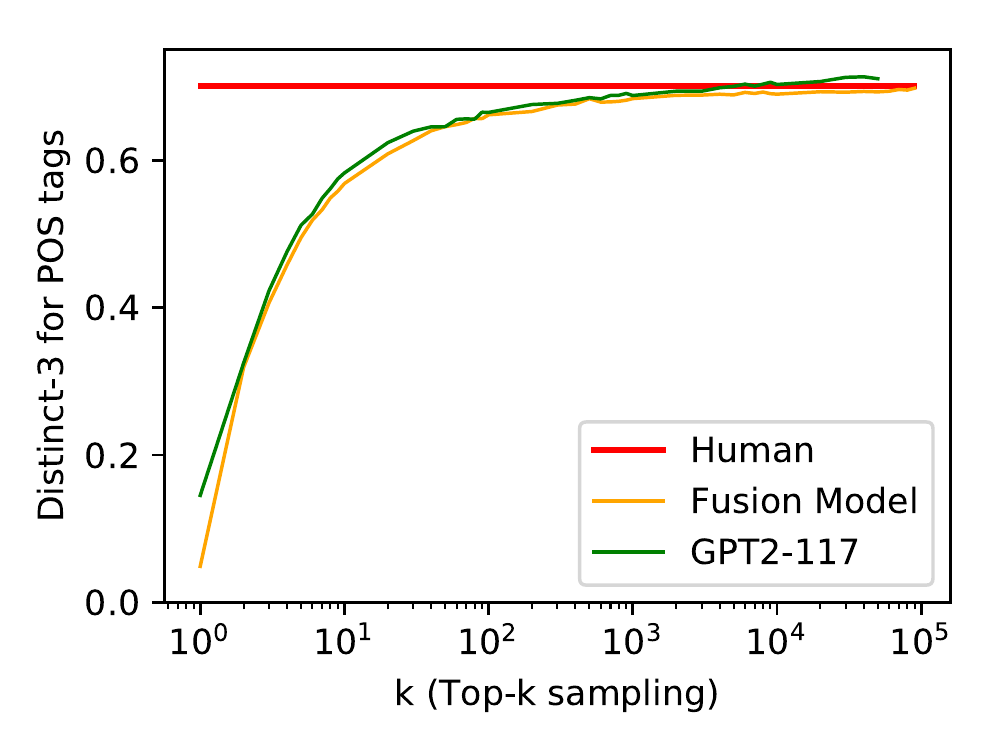}
      \caption{POS tag distinct-3 (ratio of unique POS trigrams in the story to total number of generated POS trigrams in the story).}
      \label{fig:postag_distinct3}
    \end{subfigure}
    
    \vspace{1em}
    
\caption{POS tag distinct-$n$ metric for $n=1,2,3$, for both models and all $k$. 
The ratios, which represent syntactic diversity, increase as $k$ increases, with GPT2-117 reaching human levels at $k=6000$ for unigrams, $k=9000$ for bigrams, and $k=6000$ for trigrams.
Syntactic diversity is slightly higher for GPT2-117 than for the Fusion Model for equal $k$, but the primary determining factor is $k$.
See Section \ref{sec:syn} for discussion.}
\label{fig:postag_distinctn}
\end{figure*}

\clearpage

\begin{figure*}[t!]
    \centering
    \captionsetup[subfigure]{width=0.95\textwidth,justification=raggedright}
    \begin{subfigure}[t]{0.45\textwidth}
        \centering
        \includegraphics[height=1.8in]{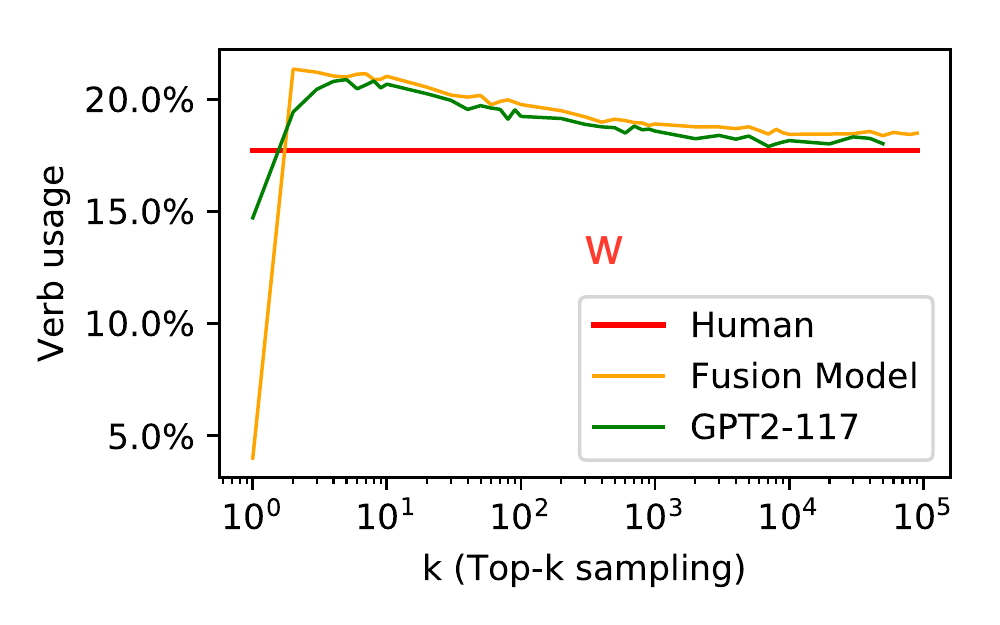}
    \end{subfigure}
    \begin{subfigure}[t]{0.45\textwidth}
        \centering
        \includegraphics[height=1.8in]{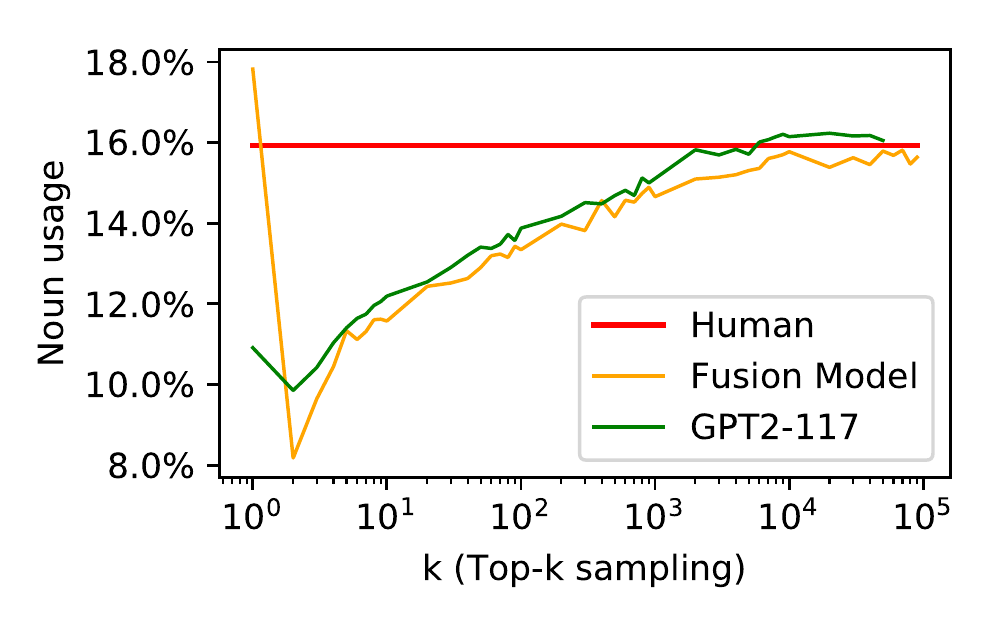}
    \end{subfigure}
    
    \begin{subfigure}[t]{0.45\textwidth}
        \centering
        \includegraphics[height=1.8in]{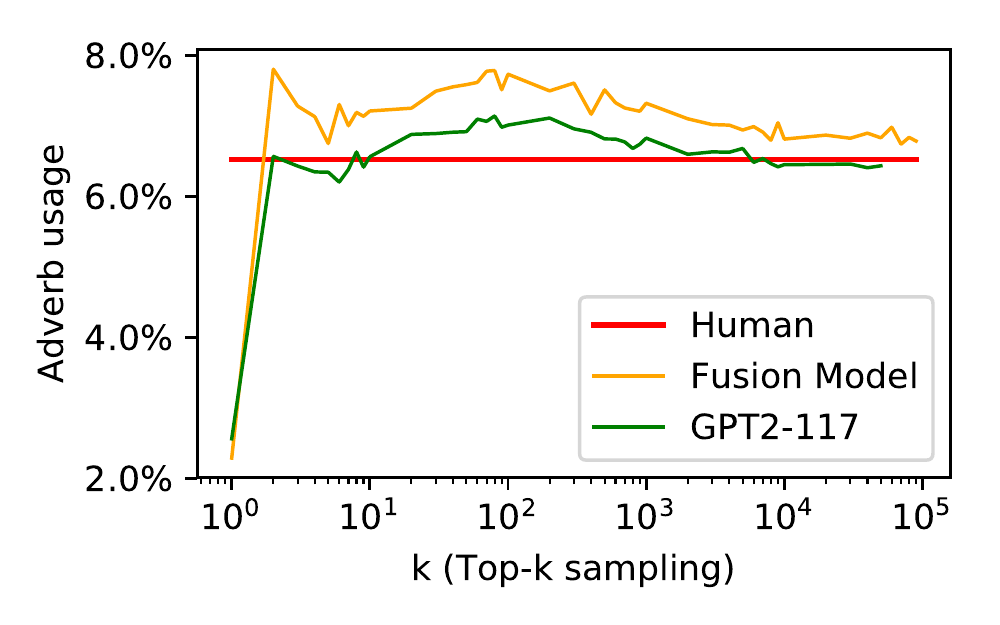}
    \end{subfigure}
    \begin{subfigure}[t]{0.45\textwidth}
        \centering
        \includegraphics[height=1.8in]{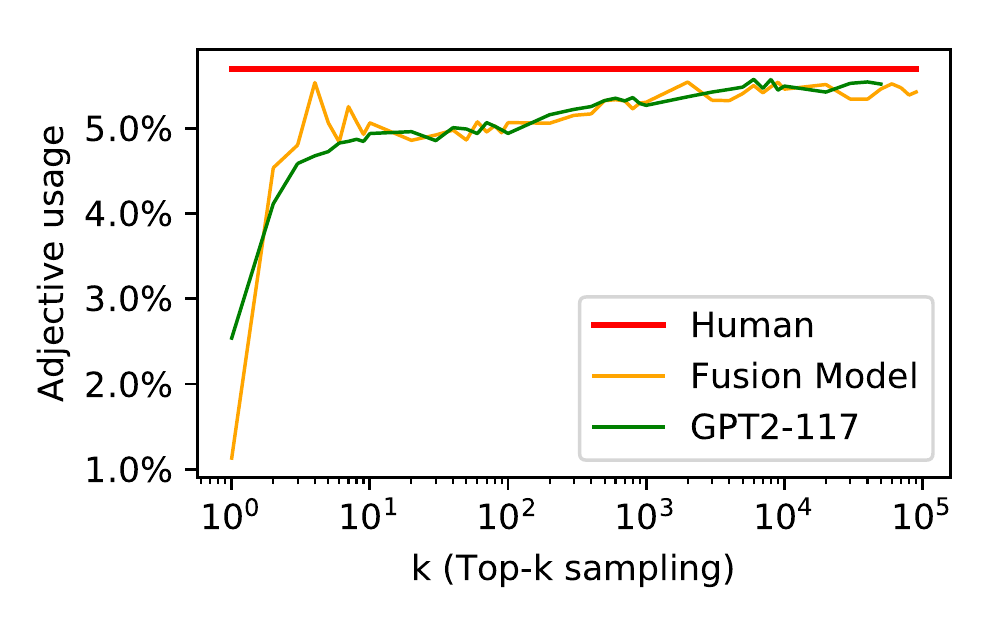}
    \end{subfigure}
    
    \begin{subfigure}[t]{0.45\textwidth}
        \centering
        \includegraphics[height=1.8in]{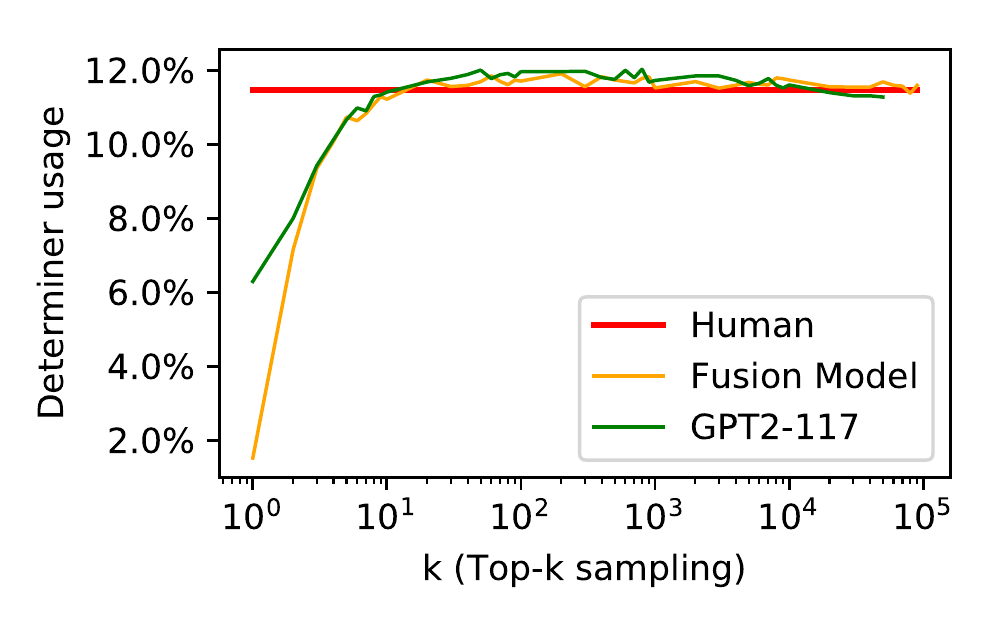}
    \end{subfigure}
    \begin{subfigure}[t]{0.45\textwidth}
        \centering
        \includegraphics[height=1.8in]{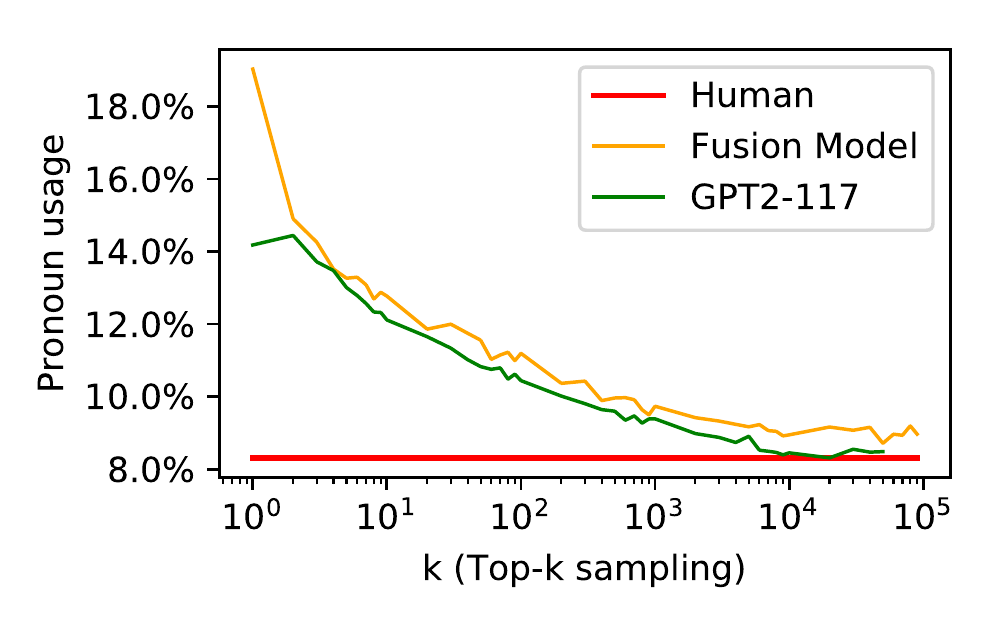}
    \end{subfigure}
    
    \begin{subfigure}[t]{0.45\textwidth}
        \centering
        \includegraphics[height=1.8in]{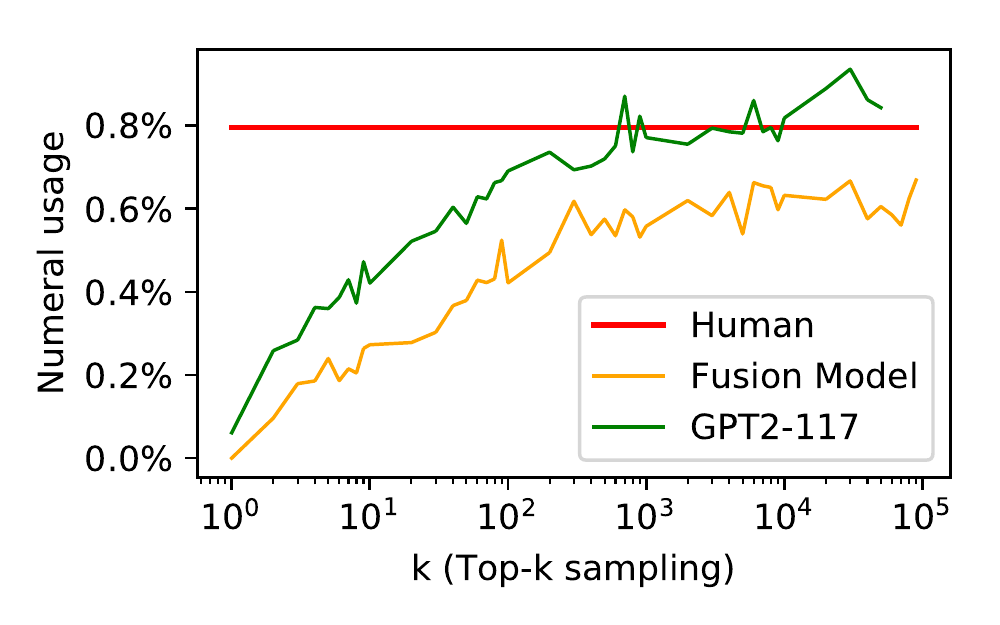}
    \end{subfigure}
    \begin{subfigure}[t]{0.45\textwidth}
        \centering
        \includegraphics[height=1.8in]{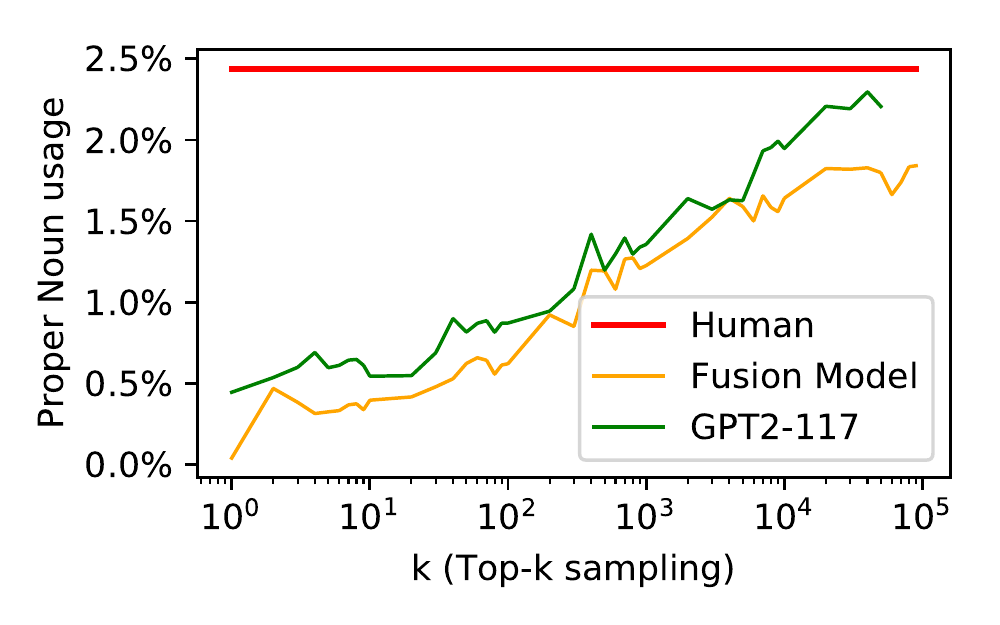}
    \end{subfigure}
    
    \caption{Usage of different POS tags in the generated stories.
    GPT2-117 tends to fit the human distribution more closely than the Fusion Model as $k$ approaches vocabulary size, in particular producing more specific POS categories such as Numeral and Proper Noun. When $k$ is small, generated text is characterized by more verbs and pronouns, and fewer nouns, adjectives, numerals and proper nouns, than human text.
    See Section \ref{sec:syn} for discussion.}
    \label{fig:postag_usage}
\end{figure*}

\clearpage

\begin{figure*}
    \begin{subfigure}{0.45\textwidth}
      \centering
      \includegraphics[height=2in]{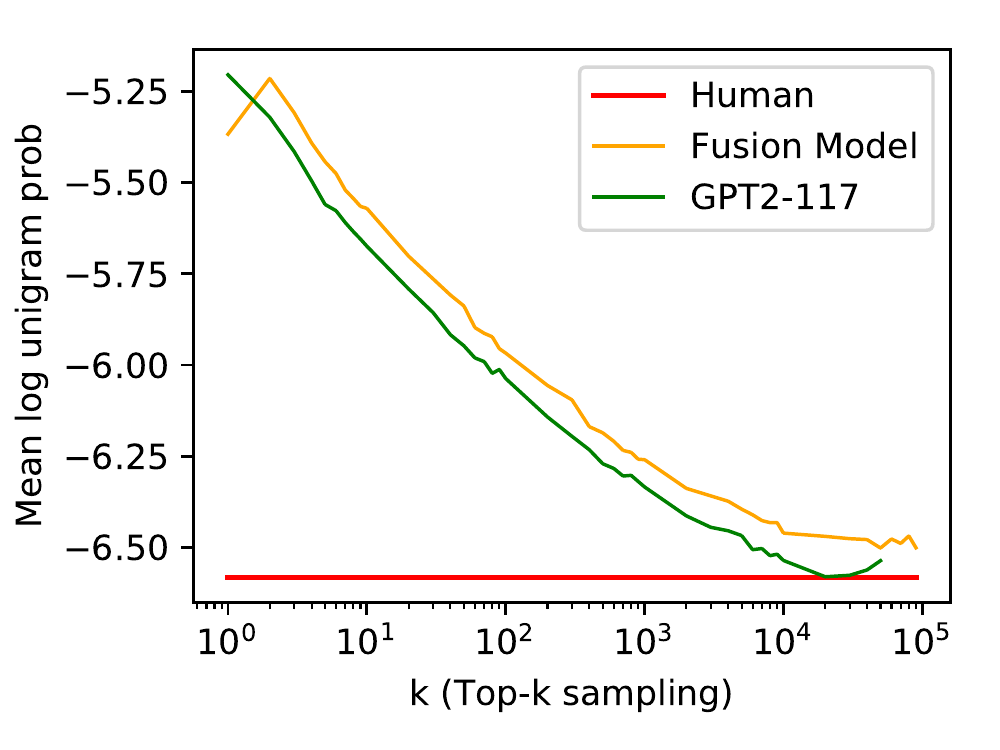}
      \caption{The mean log unigram probability of generated words.
      Higher values indicate using fewer rare words while lower values indicate using more rare words.}
      \label{fig:unigram_prob}
    \end{subfigure}
    \hspace{2em}
    \begin{subfigure}{0.45\textwidth}
      \centering
      \includegraphics[height=2in]{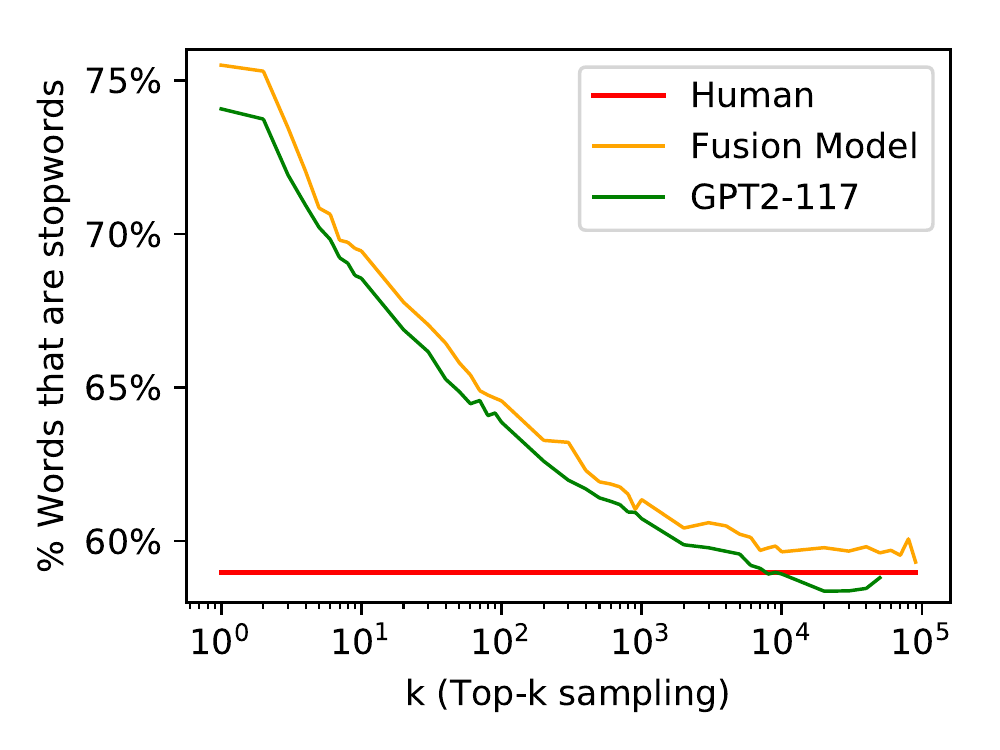}
      \caption{The percent of generated words that are stopwords, for both models, across different $k$. We use the NLTK English stopword list.}
      \label{fig:stopwords}
    \end{subfigure}
    
    \vspace{1em}
    
\caption{Rare word usage metrics for both models and all $k$.
GPT2-117 produces slightly more rare words (left) and slightly fewer stopwords (right) than the Fusion Model, for equal values of $k$.
These rareness metrics do not reach human levels until $k$ is close to vocabulary size. See Section \ref{sec:diversity} for discussion.}
\label{fig:rareness_metrics}
\end{figure*}

\begin{figure*}
    \begin{subfigure}{0.45\textwidth}
      \centering
      \includegraphics[height=2in]{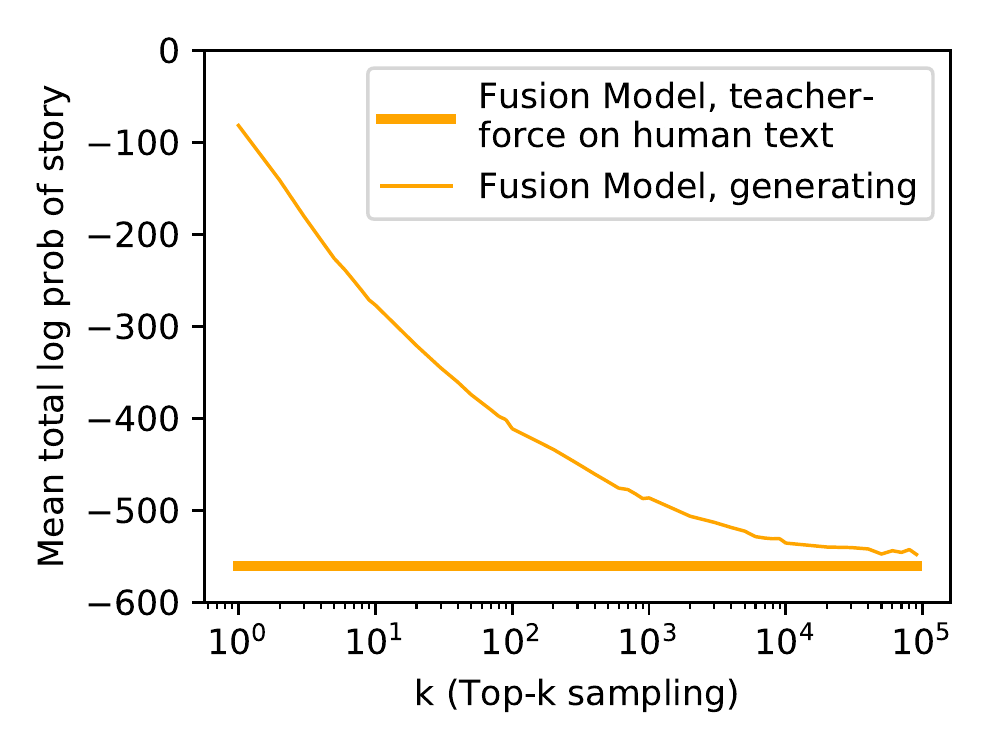}
      \label{fig:fusion_story_logprob_vs_k}
    \end{subfigure}
    \hspace{2em}
    \begin{subfigure}{0.45\textwidth}
      \centering
      \includegraphics[height=2in]{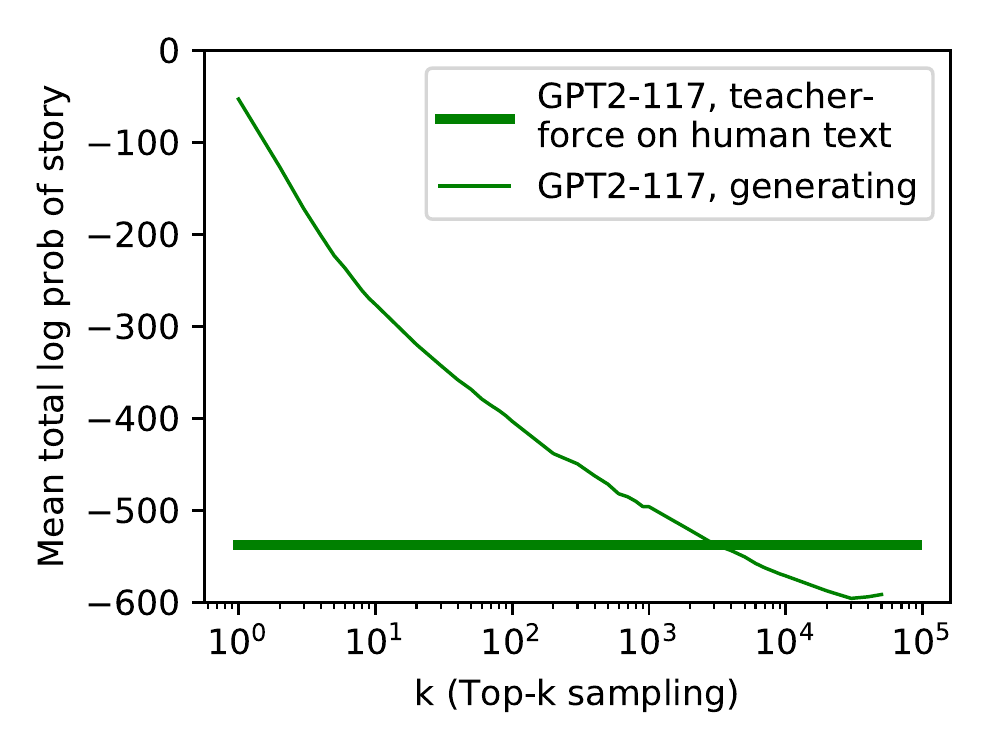}
      \label{fig:gpt2_story_logprob_vs_k}
    \end{subfigure}
\caption{The mean total log probability of the story (150 words), as measured by the models on their own generated output and on human-written stories. 
Interestingly, the Fusion Model (left) converges to the same probability it assigns to human-written stories as $k$ approaches vocabulary size, whereas GPT2-117 (right) converges to a lower probability. See Section \ref{sec:surprise} for discussion.}
\label{fig:story_logprob}
\end{figure*}

\clearpage

\begin{table*}[ht]
\small
  \centering

   \begin{tabular}{lll}
    \hline
    \multicolumn{2}{l}{\multirow{2.3}{*}{\textbf{Prompt:} A man finally discovers his superpower... well into his 80's.}}\\
    \\
    \hline 
    \multicolumn{2}{l}{} \\
    \textbf{Fusion Model ($k=10$)}: & \textbf{GPT2-117 ($k=10$)}: \\
    \multirow{7}{*}{\begin{minipage}{3in} I \verbcol{be}{34.2225} a \nouncol{child}{77.08840000000002} . \verbcol{bear}{78.85439999999998} into a \nouncol{world}{69.88959999999999} of \nouncol{darkness}{61.622499999999995} , and in the \nouncol{world}{69.88959999999999} , there \verbcol{be}{34.2225} only a few \nouncol{people}{77.7924} . My \nouncol{mother}{73.96} and I \verbcol{be}{34.2225} always alone . She \verbcol{have}{38.1924} never \verbcol{be}{34.2225} like this . But , she \verbcol{have}{38.1924} no \nouncol{idea}{31.472100000000005} what we \verbcol{be}{34.2225} together to \verbcol{be}{34.2225} . We \verbcol{be}{34.2225} always alone . We \verbcol{be}{34.2225} always together . And , of \nouncol{course}{61.15240000000001} , she \verbcol{have}{38.1924} never \verbcol{exist}{40.7044} . But , I never \verbcol{think}{41.088100000000004} about it , \verbcol{be}{34.2225}cause she \verbcol{be}{34.2225} always alone . In the \nouncol{world}{69.88959999999999} , it was n't like she \verbcol{have}{38.1924} a \nouncol{power}{36.4816} . I \verbcol{be}{34.2225} always alone , and there \verbcol{be}{34.2225} a \nouncol{time}{49.9849} when the \nouncol{light}{67.40410000000001} \verbcol{would}{26.2144} \verbcol{turn}{55.35359999999999} on . There \verbcol{be}{34.2225} a \nouncol{time}{49.9849} when I \verbcol{could}{28.5156} \verbcol{see}{51.9841} the \nouncol{light}{67.40410000000001} , and I \verbcol{could}{28.5156} \verbcol{see}{51.9841} it in her \nouncol{eye}{79.21000000000001}s , and I \verbcol{could}{28.5156} \verbcol{see}{51.9841} the \nouncol{light}{67.40410000000001} , and I \verbcol{could}{28.5156} \verbcol{see}{51.9841} it \\ \\    Mean \nouncol{noun}{35} concreteness: 3.892 \\ Mean \verbcol{verb}{35} concreteness: 2.173 \end{minipage}} & \multirow{7}{*}{\begin{minipage}{3in}A \nouncol{man}{77.26409999999998} in an old white \nouncol{coat}{80.46089999999998} \verbcol{sit}{77.44000000000001} in a \nouncol{corner}{74.1321}. He \verbcol{wear}{57.91209999999999} a black \nouncol{jacket}{78.49959999999999} with a blue \nouncol{tie}{77.61609999999997}. His \nouncol{hair}{80.46089999999998} \verbcol{be}{34.2225} grey. His \nouncol{eye}{79.21000000000001}s \verbcol{be}{34.2225} black, but his \nouncol{voice}{66.09689999999999} \verbcol{be}{34.2225} deep and clear. He \verbcol{look}{48.4416} to where the \nouncol{man}{77.26409999999998} \verbcol{be}{34.2225} \verbcol{sit}{77.44000000000001}, and he \verbcol{see}{51.9841} a \nouncol{smile}{72.25} on his \nouncol{lip}{80.28160000000001}s. It \verbcol{be}{34.2225} a \nouncol{smile}{72.25} he \verbcol{know}{32.2624} he \verbcol{would}{26.2144} \verbcol{see}{51.9841} from his own \nouncol{eye}{79.21000000000001}s. But he \verbcol{be}{34.2225} too late.   He \verbcol{be}{34.2225} on the \nouncol{sidewalk}{80.28160000000001} by the \nouncol{river}{79.03210000000001} when the \nouncol{man}{77.26409999999998} \verbcol{come}{45.15840000000001}. He \verbcol{be}{34.2225} \verbcol{wear}{57.91209999999999} a black \nouncol{coat}{80.46089999999998} with a purple \nouncol{tie}{77.61609999999997}. He \verbcol{have}{38.1924} a black \nouncol{tie}{77.61609999999997} and a white \nouncol{shirt}{79.92360000000002}. But he \verbcol{be}{34.2225} still \verbcol{wear}{57.91209999999999} a white \nouncol{suit}{80.46089999999998}. And it \verbcol{seem}{33.0625} he \verbcol{would}{26.2144} \verbcol{look}{48.4416} back at him. A \nouncol{smile}{72.25} on his \nouncol{face}{78.67690000000002}. A \verbcol{look}{48.4416} his \nouncol{friend}{49.9849} \verbcol{do}{41.7316} n't \verbcol{recognize}{36.8449}. He \verbcol{have}{38.1924} no \\ \\    Mean \nouncol{noun}{35} concreteness: 4.720 \\ Mean \verbcol{verb}{35} concreteness: 2.488\end{minipage}} \\
    \\
    \\
    \\
    \\
    \\
    \\
    \\
    \\ \\ \\ \\ \\ \\ \\ \\ \\ \\ 
    \hline
    \multicolumn{2}{l}{} \\
    \textbf{Fusion Model ($k=1000$)}: & \textbf{GPT2-117 ($k=1000$)}: \\
    \multirow{7}{*}{\begin{minipage}{3in}For a brief \nouncol{moment}{31.472100000000005}, the dark \nouncol{green}{65.12490000000001} of his \nouncol{eye}{79.21000000000001}s \verbcol{flash}{58.8289} the last \nouncol{line}{72.25}s of \nouncol{reality}{32.718399999999995} into \nouncol{existence}{30.6916}, finally \verbcol{fade}{41.99040000000001} slowly into \nouncol{existence}{30.6916} so that he \verbcol{could}{28.5156} not \verbcol{feel}{39.438399999999994} it. Only the \nouncol{million}{47.61000000000001}s of \nouncol{pixel}{62.884899999999995}s \verbcol{float}{65.12490000000001} in his peripheral \nouncol{vision}{54.612100000000005} almost \verbcol{disappear}{47.05959999999999}. His \nouncol{radio}{76.3876} \verbcol{respond}{44.089600000000004} to the pinging very hard \nouncol{silence}{53.7289} of the outside \nouncol{world}{69.88959999999999}. Seven \nouncol{people}{77.7924} \verbcol{have}{38.1924} \verbcol{visit}{62.7264} his main internal \nouncol{network}{51.84} for what \verbcol{seem}{33.0625} like a \nouncol{lifetime}{37.822500000000005} but this \nouncol{time}{49.9849}, the only \nouncol{switch}{65.12490000000001} to an external \nouncol{supply}{55.204899999999995} \nouncol{system}{48.163599999999995} that he \verbcol{could}{28.5156} simply \verbcol{take}{49.84360000000001} \nouncol{advantage}{36.0} of. Unable to \verbcol{convey}{32.718399999999995} \nouncol{feeling}{32.2624}s about the last \nouncol{word}{57.153600000000004}s he \verbcol{would}{26.2144} \verbcol{have}{38.1924} to \verbcol{endure}{36.0}, but it \verbcol{have}{38.1924} \verbcol{respond}{44.089600000000004} to the innumerable \nouncol{message}{63.52090000000001}s and countless sleepless \nouncol{hour}{50.41}s. Most of them \verbcol{be}{34.2225} always available on its \nouncol{surface}{68.2276}, just to \verbcol{make}{44.4889} sure.   In his \nouncol{quest}{47.1969} for to \verbcol{spend}{48.024899999999995} \nouncol{eternity}{29.8116} on \nouncol{death}{59.752900000000004}, he \verbcol{send}{44.89} \\ \\    Mean \nouncol{noun}{35} concreteness: 3.201 \\ Mean \verbcol{verb}{35} concreteness: 2.435 \end{minipage}} & \multirow{7}{*}{\begin{minipage}{3in}( First \nouncol{time}{49.9849} \nouncol{poster}{74.9956}, \verbcol{hope}{27.5625} its \nouncol{ok}{37.33209999999999} )   The young \nouncol{boy}{76.7376}, \verbcol{watch}{74.1321} \nouncol{tv}{81.0}, \verbcol{spot}{67.40410000000001} the \nouncol{television}{77.9689} \nouncol{onscreen}{55.0564}, before \verbcol{glance}{55.95040000000001} around to \verbcol{see}{51.9841} the \nouncol{screen}{73.96} \verbcol{start}{45.0241} the \nouncol{countdown}{53.29} on the \nouncol{tv}{81.0}, \verbcol{point}{54.612100000000005} to the \nouncol{screen}{73.96} in `` It 's both the same. ''   ``... \verbcol{let}{39.187599999999996} 's... \verbcol{let}{39.187599999999996} 's \verbcol{try}{38.68840000000001} this and... we \verbcol{will}{44.089600000000004} \verbcol{team}{60.6841} up so that... we \verbcol{can}{73.1025}... \verbcol{have}{38.1924} the same \nouncol{power}{36.4816}....like... so we \verbcol{can}{73.1025} \verbcol{use}{45.96839999999999} this \nouncol{superpower}{47.1969} over and over again. ''   A brief \nouncol{silence}{53.7289}. Only a familiar \nouncol{conversation}{52.99839999999999}, \verbcol{interrupt}{49.0} his mad \nouncol{dash}{54.612100000000005} \nouncol{movement}{58.216899999999995}, \verbcol{follow}{52.7076} with his high \verbcol{pitch}{64.0} slurred and \verbcol{wither}{45.0241} \nouncol{voice}{66.09689999999999} :   `` I ca n't \verbcol{stand}{66.5856} \nouncol{anyone}{44.89} \verbcol{talk}{65.12490000000001} like that son*s*. ''   More casual \nouncol{conversation}{52.99839999999999} that \verbcol{interrupt}{49.0} his childish \nouncol{step}{72.93159999999999} \verbcol{be}{34.2225} \verbcol{rush}{47.61000000000001} to the \nouncol{scissor}{79.21000000000001}s. \\ \\ \\ \\ \\ \\    Mean \nouncol{noun}{35} concreteness: 3.793 \\ Mean \verbcol{verb}{35} concreteness: 3.162 \end{minipage}} \\

    \\
    \\
    \\
    \\
    \\
    \\
    \\
    \\
    \\
    \\
    \\
    \\
    \\ \\ \\ \\ \\ \\ \\ \\ \\
    \hline
    \end{tabular}
\caption{Generated stories from both models, under $k=10$ and $k=1000$. 
Nouns are highlighted in green and verbs in yellow. 
The highlighting intensity reflects the word's concreteness rating.
For equal $k$, GPT2-117 generally generates more concrete words than the Fusion Model.
For both models, low $k$ is characterized by high noun concreteness (e.g. physical objects such as \textit{jacket}) and low verb concreteness (e.g. non-physical actions such as \textit{be}).
Conversely, high $k$ is characterized by low noun concreteness (e.g. abstract concepts such as \textit{reality}) and high verb concreteness (e.g. physical actions such as \textit{talk}).
See Section \ref{sec:concrete} for discussion.}
\label{fig:concrete_ex}
\end{table*}

\clearpage

\begin{figure*}
    \begin{subfigure}{0.45\textwidth}
      \centering
      \includegraphics[height=2in]{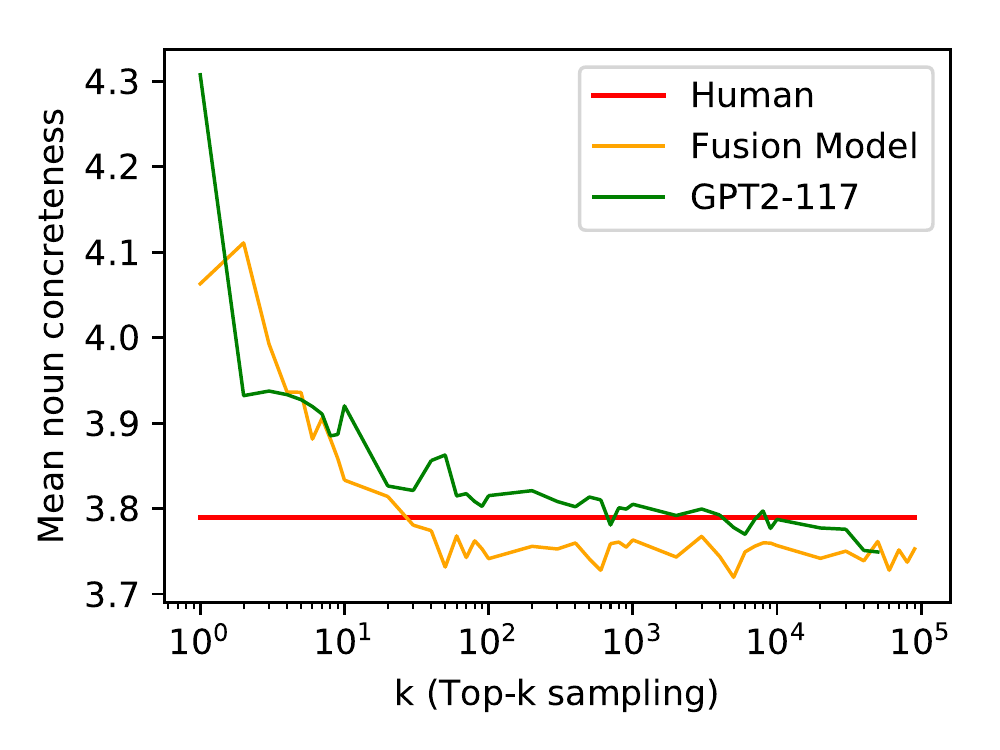}
      \caption{Mean concreteness rating (1-5) of nouns in the story.}
      \label{fig:concreteness_noun}
    \end{subfigure}
    \hspace{2em}
    \begin{subfigure}{0.45\textwidth}
      \centering
      \includegraphics[height=2in]{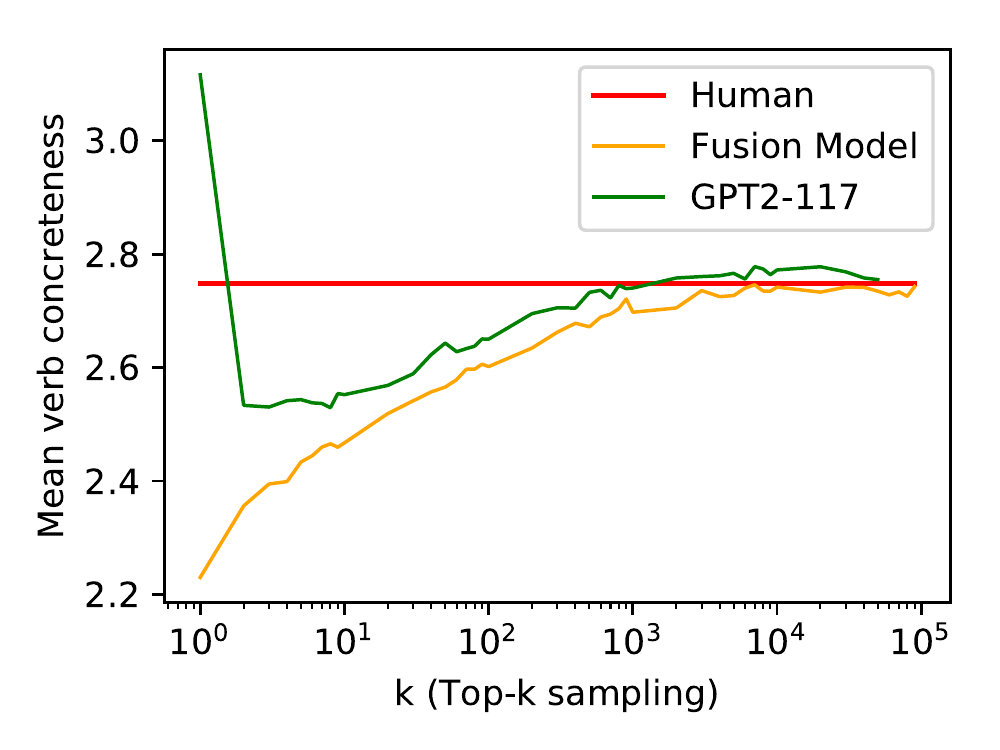}
      \caption{Mean concreteness rating (1-5) of verbs in the story.}
      \label{fig:concreteness_verb}
    \end{subfigure}
    
    \vspace{1em}
    
\caption{Mean concreteness rating of the nouns and verbs in the story, for both models and all $k$. 
GPT2-117 generally produces nouns and verbs that are more concrete than the Fusion Model for the same $k$.
For both models, as $k$ increases, noun concreteness reduces and verb concreteness increases.
See Section \ref{sec:concrete} for discussion.}
\label{fig:concreteness}
\end{figure*}

\begin{table*}[b]
\small
  \centering
  \begin{tabular}{l}
    \hline \\
    \begin{minipage}{6in} \begin{flushleft}\textbf{Prompt:} \textit{In an alternative reality where sleep is non-existent among living beings, our protagonist (spontaneously or after an event) falls asleep in which he/she experiences for the first time in human history what a dream is.} \end{flushleft}\end{minipage} \\ \\ \hline \\
    \begin{minipage}{6in} \begin{flushleft}\textbf{GPT2-117 ($k=1000$):} \textit{I sat in my bed as my girlfriend sat behind me, buzzing into her e-reader, letting the day's stories write themselves on her's monitor. Like a blur, all the usual high-asyllabic drivel and senseless ramblings that normally attracted the attention of a horrid sleeping creature huddled about me like a faucet. She did not know how I placed this car with her. But I tried, first tried to ignore that I had hired the services of a dog to help and then quietly used it in a desperate bid to drive the car through the lawn. Each and every day, I watched her drool down an old dusty hardwood mattress her beady eyes trying desperately to think of this rotting dream.} [...] \end{flushleft}\end{minipage} \\ \\ \hline
\end{tabular}
\caption{An example of syntactically natural but nonsensical text, generated with high $k$. Though the text is fluent and readable, it is surreal and bizarre. See Section \ref{sec:conclusion} for discussion.}
\label{tab:weird}
\end{table*}

\end{document}